%% file: ide_et_al.tex
\documentclass[conference]{IEEEtran}
\IEEEoverridecommandlockouts
\usepackage{cite}
\usepackage{amsmath,amssymb,amsfonts}
\usepackage{algorithmic}
\usepackage{graphicx}
\usepackage{textcomp}
\usepackage{xcolor}

\usepackage{widetext}

\def\BibTeX{{\rm B\kern-.05em{\sc i\kern-.025em b}\kern-.08em
    T\kern-.1667em\lower.7ex\hbox{E}\kern-.125emX}}

\usepackage{booktabs}       
\usepackage{amsfonts}       
\input{ide_preamble}

\begin{document}

\title{Decentralized Collaborative Learning with Probabilistic Data Protection}

\author{
\IEEEauthorblockN{Tsuyoshi~Id\'e}
\IEEEauthorblockA{\textit{IBM Research, T. J. Watson Research Center} \\
Yorktown Heights, NY, USA \\
tide@us.ibm.com}
\and
\IEEEauthorblockN{Rudy Raymond}
\IEEEauthorblockA{\textit{IBM Research -- Tokyo} \\
Tokyo, Japan \\
rudyhar@jp.ibm.com}
}


\maketitle

\begin{abstract}
We discuss future directions of Blockchain as a collaborative value co-creation platform, in which network participants can gain extra insights that cannot be accessed when disconnected from the others. As such, we propose a decentralized machine learning framework that is carefully designed to respect the values of democracy, diversity, and privacy. Specifically, we propose a federated multi-task learning framework that integrates a privacy-preserving dynamic consensus algorithm. We show that a specific network topology called the expander graph dramatically improves the scalability of global consensus building. We conclude the paper by making some remarks on open problems.\footnote{\textcolor{blue}{Published as: Tsuyoshi Id\'e and Rudy Raymond, ``Decentralized Collaborative Learning with Probabilistic Data Protection,'' Proceedings of the 2021 IEEE International Conference on Smart Data Services (SMDS 21, September \\ 5-10, 2021, virtual), pp.234-243.}}
\end{abstract}


\section{Introduction} \label{sec:Introduction}

Spurred by the remarkable commercial success of cryptocurrencies, there have been many attempts to date to extend Blockchain as a decentralized management platform for general business transactions. In most application scenarios, such as product traceability~\cite{tse2017blockchain,lu2017adaptable,toyoda2017novel} and those involving smart contracts~\cite{christidis2016blockchains}, Blockchain has been used essentially as an immutable data storage whose goal is simply to manage identical replicas of data among distributed nodes. Although this is a meaningful first step, we argue that the true value of Blockchain lies in its potential for value co-creation by network participants (``agents'') through knowledge sharing~\cite{scekic2018blockchain,seebacher2017blockchain,kondrateva2021potential}. The platform should help the agents obtain extra insights that cannot be accessed when looking at their own data alone, disconnected from the others. We envision that the next generation of Blockchain will integrate \textit{collaborative learning} capabilities at the core. 

Inspired by the Nakamoto's original agenda~\cite{nakamoto2008bitcoin}, our collaborative learning platform features three main characteristics: 1) The entire learning procedure is done in a decentralized manner, i.e.,~without relying on the central authority. 2) The outcomes of the learning reflect specific situations of the individual agents, resulting in generally different models for each of the agents. 3) The data and the learned model of the agents are protected as a private property. 

\begin{figure}
\centering
    \includegraphics[trim={3cm 0.5cm 4cm 0cm},clip,height=4.5cm]{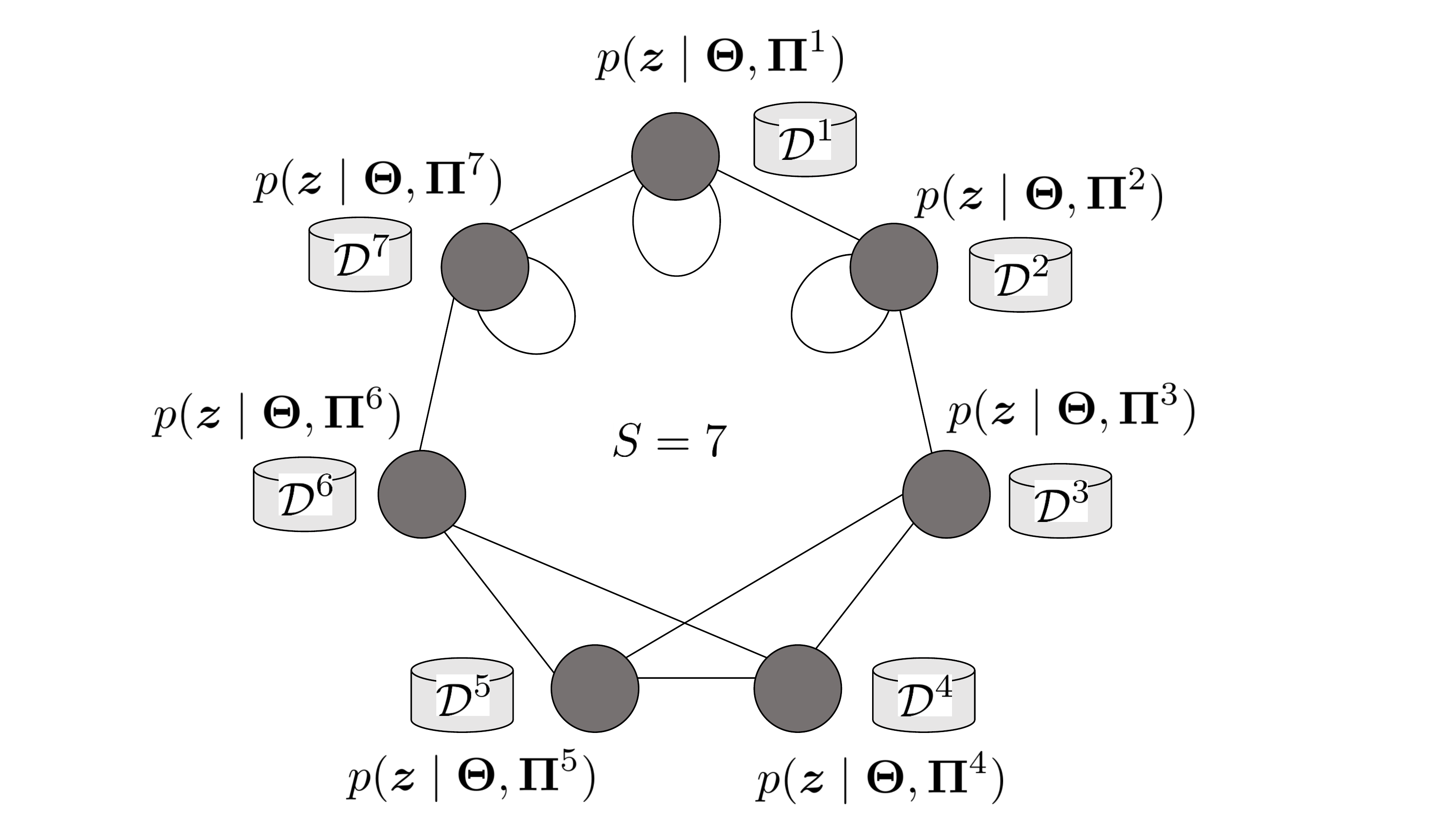}
  \caption{Illustration of decentralized multi-task learning illustrated on a 3-regular expander graph of $S=7$ nodes, where $p(\cdot \mid \bm{\Theta},\bm{\Pi}^s)$ denotes a statistical machine learning model for the agent $s$ with $\bm{\Theta}$ being common model parameters and $\bm{\Pi}^s$ being agent-specific model parameters. }
\label{fig:S7expander_model.pdf}
\end{figure}

The \textit{first} characteristic is to ensure \textit{democracy} in the platform. The agents voluntarily communicate with the others through given network infrastructure, but there is no such thing as the central server that collects a piece of data from the agents and perform, e.g.,~stochastic gradient descent to train a deep learning model. The \textit{second} characteristic is to ensure \textit{diversity} in the platform. Here, a `model' refers to a probability distribution in general that captures patterns hidden in the data, as implied in Fig.~\ref{fig:S7expander_model.pdf}. In the machine learning literature, this is commonly called \textit{multi-task learning} because the goal is to learn $S$ different models (`tasks') simultaneously, where $S$ is the number of agents participated in the network. The \textit{third} characteristic is to ensure the \textit{privacy} of the agents in the platform. Although they learn from the other agents and share their learnings in return in a certain way, their original data and the resulting model must be protected.

This paper proposes a new framework of decentralized collaborative learning with certain privacy guarantees. Given a specific parametric model agreed upon, the goal of the agents is to maximize the likelihood function of the entire system in a collaborative manner. Higher likelihood implies higher model fidelity, which will lead to better business outcomes through more accurate predictions. In our previous work~\cite{ide2019efficient}, we presented a decentralized multi-task learning protocol for a mixture of exponential family distributions, where the dynamic consensus algorithm~\cite{wu2005agreement,ren2005survey,olfati2007consensus} eliminates the need for the central server. We found that the spectral structure of network topology plays a critical role in the convergence of the algorithm, but the analysis was mainly on the cycle graph, in which an analytic form of the eigenspectrum is available. In this paper, we show that a family of graphs called the \textit{expander graph}~\cite{vadhan2012pseudorandomness,lubotzky2010discrete} can dramatically speedup global consensus. We also provide a systematic analysis on the probability of privacy breach in a couple of different scenarios. Decentralized collaborative learning is a new research field. We conclude the paper with some remarks on open problems.

\section{Related work}

One recent major trend in distributed learning is to distributedly train deep neural networks using SGD (stochastic gradient descent)~\cite{mohassel2017secureml,mcmahan2017communication,Konecny16NIPSworkshop,agarwal2018cpsgd,yang2019federated}, 
in which managing huge computational overhead has been an issue. Although most of the existing work falls into the category of distributed \textit{single}-task learning, some recent works point out that model biases and thus a lack of diversity can be a serious issue~\cite{xu2014distributed,Mohri19ICML}. A multi-task extension is discussed in~\cite{smith2017federated} without a data privacy context. A unique role of the exponential family in data privacy is pointed out in~\cite{bernstein2018differentially}.

For privacy preservation in distributed learning, a common approach is to use differential privacy~\cite{xie2017privacy,heikkila2017differentially}, but for real-valued data, performance degradation in learning due to introduced noise is an open question, as discussed in~\cite{ding2018comparing}. Privacy preservation in a decentralized environment is a challenging task in general. Almost the only solution known so far is the use of homomorphic encryption and its variants~\cite{mohassel2017secureml,liu2020secure,ruan2017secure}, but its computational cost is known to be often prohibitive.

\section{Multi-Task learning framework } \label{sec:framework}

This section summarizes the problem setting from our previous work~\cite{ide2019efficient}.

\subsection{Problem setup} \label{subsec:setting}

As shown in Fig.~\ref{fig:S7expander_model.pdf}, there are $S \geq 3$ agents in the network. Each agent indexed by $a$ privately keeps its own data set 
\begin{align}
    \mathcal{D}^a \triangleq \{\bmz^{(1)},\ldots, \bmz^{(N^a)} \}
\end{align}
about a random variable $\bmz$. Here, $N^a$ is the number of samples of the $a$-th agent. As a general rule, we use the superscript such as $^{(n)}$ to represent the $n$-th instance of a random variable. The random variable $\bmz$ can be a pair of a feature vector and its label like $\bmz=(y,\bmx)$ or simply a feature vector alone $\bmz = \bmx$. We assume $\bmx$ is \textit{real-valued} and noisy in general. The goal of the agents is to learn a predictive distribution for~$\bmz$. As illustrated in Fig.~\ref{fig:S7expander_model.pdf}, the distribution will have model parameters $\bm{\Theta}$ and $\bm{\Pi}^a$, where $\bm{\Theta}$ is parameters shared by all the agents and $\bm{\Pi}^a$ is agent-specific parameters. Both $\bm{\Theta}$ and $\bm{\Pi}^a$ are to be learned from the total data $\mathcal{D} \triangleq\{\mathcal{D}^1,\ldots, \mathcal{D}^S \}$.  The question is how the agents leverage information from the other ones while keeping data privacy. 

As part of the network infrastructure, a set of bi-directional communication paths are given as an undirected graph, whose nodes are the agents and the edges are pairwise communication paths, as shown in the figure. As is the case in the IP (internet protocol) network of the Internet, the infrastructure is assumed to do basic bookkeeping jobs such as network routing and clock synchronization without any interest in the contents communicated. Network failures are unavoidable in real networks but we do not consider them for simplicity. We also assume a consortium-based network, where the agents have verified identities. The agents are \textit{honest but curious}, meaning that they do not lie about computed statistics but they always try to selfishly get as much information as possible from the other agents. 


\subsection{Mixture of exponential family} \label{subsec:probmodel}

To capture the diversity among the agent, we employ a \textit{mixture} of the exponential family. In what follows, we focus on the case where $\bmz=\bmx \in \mathbb{R}^M$ with $M$ being the dimensionality of the variable and the learning task is unsupervised (a.k.a.~density estimation). Extension to the supervised setting can be done easily. Practical applications of this setting include failure detection of industrial robots, where all the $S$ agents are assumed to have the same set of physical sensors. See~\cite{ide2018collaborative} for more detail. 

Now the observation model of the $a$-th agent is given by:
\begin{align}\label{eq:mixtureModel}
p(\bmx \mid \bm{\Theta}, \bm{u}^a) &=
\prod_{k=1}^K f(\bmx \mid {\bm \theta}_k)^{u_k^a} \\
\label{eq:exponentialFamilyDistribution}
f(\bmx \mid \bm{\theta}_k) &= G(\bm{\theta}_k)H(\bmx)\exp\{ \bm{\eta}( \bm{\theta}_k)^\top \bm{T}(\bmx)\},
\end{align}
where $K$ is the number of mixture components and $\bmu^a \triangleq (u^a_1,\ldots,u^a_K)^\top$ is the one-hot indicator variable representing cluster assignment. Notice that the dependency on $a$ in $\bmu^a$ represents diversity of the model. In $f(\bmz \mid \bm{\theta}_k)$, functional forms of $G, H$ (scalar function) and $\bm{\eta},\bm{T}$ (vector-valued function) are given by a specific choice in the exponential family~\cite{betulehmann2006theory}. We will give a Gaussian-based model as an example below.

In the unsupervised scenario, the observation model in the form Eq.~\eqref{eq:mixtureModel} is almost always combined with a prior distribution of $\bmu^a$: 
\begin{align}\label{eq:prior_pi}
p({\bm u}^a \mid {\bm \pi}^a) &= \mathrm{Cat}({\bm u}^a \mid {\bm \pi}^a) 
\triangleq \prod_{k=1}^K (\pi_k^a)^{u^a_k}
\end{align}
where $\mathrm{Cat}$ denotes the categorical distribution. The parameter $\bmpi^a \triangleq (\pi^a_1,\ldots, \pi^a_K)$ can be viewed as the probability distribution over the $K$ clusters and satisfies $\sum_{k=1}^K \pi^a_k =1$. 

For stable numerical estimation, prior distributions are imposed also on $\bm{\Theta} \triangleq \{ \bmtheta_k\}$ and $\bm{\Pi} \triangleq \{ \bmpi^a\}$ as
\begin{align}
    p(\bm{\Theta}) &= \prod_{k=1}^K p(\bm{\theta}_k), \quad 
    p({\bm \Pi}) = \prod_{a=1}^S p({\bm \pi}^a),
\end{align}
where we have used $p(\cdot)$ to generically represent potentially different probability distributions. 

Here we give an example when $f(\bm{z} \mid \bm{\theta}_k)$ is the Gaussian $\mathcal{N}(\bm{x} \mid \bm{\mu}_k,(\mathsf{\Lambda}_k)^{-1})$, where $\bm{\mu}_k$ is the mean and $\mathsf{\Lambda}_k$ the precision matrix. For $\bm{\theta}_k = \{ \bm{\mu}_k, \mathsf{\Lambda}_k \}$, one practically recommended choice for the prior distribution is the Gauss-Laplace distribution:
\begin{gather}\label{eq:Gauss-Laplace}
p({\bm \mu}_k,{\sf \Lambda}_k) \propto
{\cal N}({\bm \mu}_k|{\bm m}_0,(\lambda_0{\sf \Lambda}_k)^{-1})
\exp\left( -\frac{\rho}{2} \| {\sf \Lambda}_k \|_1 \right)
\end{gather}
where $\rho, \lambda_0,\bm{m}_0$ are predefined constants and $\| \cdot \|_1$ is the $\ell_1$ norm. For $p({\bm \pi}^a)$, a common choice is the Dirichlet distribution $p(\bm{\pi}^a)\propto (\pi^a_1\cdots\pi^a_K)^\gamma$ with $\gamma \sim 1$ is a given constant.

\subsection{Model estimation algorithm} \label{subsec:MAP}
 
The probabilistic model presented above contains unknown model parameters $\bm{\Theta},\bm{\Pi}$ in addition to the latent variable $\bmU \triangleq \{ \bmu^1, \ldots, \bmu^S\}$. The standard strategy to learn the model in such a case is to maximize the log marginalized likelihood $L_0$ with respect to $\bm{\Theta},\bm{\Pi}$:
\begin{align} \nonumber 
L_0&\triangleq
\ln p({\bm \Pi})p(\bm{\Theta}) \\ \nonumber
&+ \ln \left\{\sum_{\bm{U}}  
\prod_{a=1}^s\prod_{n=1}^{N^s}p(\bm{z}^{a(n)}| \bm{\Theta}, \bm{u}^{a(n)})p(\bmu^{a(n)} \mid \bmpi^a)
\right\}.
\end{align}
Unfortunately, this maximization problem is intractable even in the simple Gaussian case. We instead maximize the lower bound of $L_0$ derived by applying Jensen's inequality. We can derive a simple iterative algorithm to estimate the unknown model parameters $\bm{\Theta},\bm{\Pi}$ as well as the posterior distribution of the latent  variable $\bmU$:
\begin{align}
    Q(\bmU) = \prod_{a=1}^S\prod_{n=1}^{N^s}\prod_{k=1}^K (r^{a(n)}_k)^{u^{a(n)}_k}.
\end{align}

Here, we summarize the result presented in our previous work~\cite{ide2019efficient}. With an initialized $\{r^{a(n)}\}$ so $\sum_{k=1}^K r^{a(n)}_k = 1$, we repeat the following steps until convergence:
\begin{itemize}
    \item In each $a \in \{1,\ldots,S\}$, with the latest $\{r_k^{a(n)}\}$, locally compute 
    \begin{align}\label{eq:localStats}
        N_k^a \triangleq \sum_{n=1}^{N^a} r^{a(n)}_k  \;\;\mbox{and} \;\;
\bm{T}^a_k \triangleq \sum_{n=1}^{N^a} r^{a(n)}_k \bm{T}(\bm{z}^{a(n)}). 
    \end{align}
    \item Among all $a=1,\ldots, S$, build a global consensus on
    \begin{align}\label{eq:aggregate_global_exponential_family}
        N_k \triangleq \sum_{a=1}^{S} N^{a}_k,  \quad \mbox{and} \quad
\bm{T}_k \triangleq \sum_{a=1}^{S} \bm{T}^a_k,
    \end{align}
     \item In each $a \in \{1,\ldots,S\}$, locally solve
     \begin{align} \label{eq:solve}
         \max_{\bm{\theta}_k} \left\{
\ln p(\bm{\theta}_k)+N_k \ln G(\bm{\theta}_k) + \bm{T}_k^\top\bm{\eta}(\bm{\theta}_k) 
\right\}
\end{align}
 \item In each $a \in \{1,\ldots,S\}$, with the latest $\{\bmtheta_k\}$ and $ \{\bmpi^a\}$ with $\pi^a_k = \frac{N_k^a +\gamma}{N^a +K\gamma}$, locally update
\begin{align}\label{eq:QofZ}
     r_k^{a(n)}= \frac{
\pi_k^a f(\bm{z}^{a(n)} \mid \bm{\theta}_k) 
}{
\sum_{m=1}^K \pi_m^a f(\bm{z}^{a(n)} \mid \bm{\theta}_m) 
}.
\end{align}
\end{itemize}
Here, we assumed that we have used the Dirichlet distribution $p(\bm{\pi}^a)\propto (\pi^a_1\cdots\pi^a_K)^\gamma$ for $p({\bm \Pi})$. Notice that agent-agent communication is involved only in the second step; All the other steps need only local computation that can be complete within each agent. The complexity per agent per iteration is $\calO(N^a + M^3 + \ln S)$, assuming Eq.~\eqref{eq:solve} takes $M^3$ and the consensus step Eq.~\eqref{eq:aggregate_global_exponential_family} takes $\ln S$ (See Theorem~\ref{prop:expander}).

\subsection{Gaussian example}
\label{app:GaussianExample}

To be concrete, we provide parameter updating equations for the Gaussian observation model with the Gauss-Laplace prior Eq.~\eqref{eq:Gauss-Laplace}. In this case, instead of $\bmT^a_k$ in Eq.~\eqref{eq:localStats}, we compute 
\begin{align}
\bm{m}^a_k = \sum_{n=1}^{N^s}r_k^{a(n)}\bm{x}^{a(n)}, \;\;\;
\mathsf{C}^a_k = \sum_{n=1}^{N^s}r_k^{a(n)}\bm{x}^{a(n)}{\bm{x}^{a(n)}}^\top.
\end{align}
for each $a\in\{1,\ldots, S\}$ and each mixture component $k\in\{1,\ldots,K\}$. Then, in the step of global consensus in Eq.~\eqref{eq:aggregate_global_exponential_family}, we compute aggregated values as
\begin{align}\label{eq:Gaussian-consensus}
\bar{\bm{m}}_k  = \sum_{a=1}^S \bm{m}^a_k,
\;\;\;
\bar{\mathsf{C}}_k  = \sum_{a=1}^S \mathsf{C}^a_k.
\end{align}
Finally, in the step of optimization in Eq.~\eqref{eq:solve}, we compute
\begin{gather}
\label{eq:Nk}
\bm{\mu}_k = \frac{1}{\lambda_0 + N_k}\bar{\bm{m}}_k,\;\;\;
\mathsf{\Sigma}_k =  \frac{1}{N_k}\bar{\mathsf{C}}_k + \bm{\mu}_k \bm{\mu}_k^\top,
\\
{\sf \Lambda}_k = \arg
\max_{{\sf \Lambda}_k}\left\{
\ln \det{\sf \Lambda}_k
- \mathrm{Tr}({\sf \Lambda}_k\mathsf{\Sigma}_k) -\frac{\rho}{N_k}\|{\sf \Lambda}_k \|_1 \right\},
\end{gather}
where $\mathrm{Tr}$ is the matrix trace and $\det$ is the matrix determinant. This optimization problem is well-known in covariance selection and can be solved very efficiently with the graphical lasso algorithm~\cite{Friedman08glasso,hsieh2014quic}. To ensure that all the agents have the same $\{\bmmu_k,{\sf \Lambda}_k\}$, they can run another global consensus step to register the average as the final outcome in each iteration round.

\section{Aggregation via secret sharing}\label{sec:Aggregation}

This section focuses on Eq.~\eqref{eq:aggregate_global_exponential_family}, i.e.,~how to securely aggregate the local statistics. Since aggregation can be done element-wise, without loss of generality, we consider the problem of computing the sum of scalars $\{ \xi_a \}$: 
\begin{align}\label{eq:averageConsensus}
\bar{\xi} = \sum_{a=1}^S \xi_a = \bm{1}_S^\top\bm{\xi}(0), 
\end{align}
where $\xi_a$'s are constants to be summed, $\bm{1}_S$ is the $S$-dimensional vector of ones, and we defined $\xi_a(0) = \xi_a$. Aggregation would be trivial if a trusted coordinator existed in the network. The question is how to compute the summation only through local communications and how to make it secure.

\subsection{Aggregation through Markov transitions on graph}\label{subsec:dynamicalConsensus}

Let $\sfA \in \{0,1\}^{S\times S}$ be the incidence matrix of the communication graph, where only connected nodes (or neighboring nodes) can communicate with each other. As illustrated in Fig.~\ref{fig:S7expander_model.pdf}, the graph is undirected but may have self-loops and multiple edges. Given $\sfA=[A_{a,j}]$, consider the following updates:
\begin{align}\label{eq:Dynamics_elementwise}
\xi_a(t+1) = \xi_a(t) + \epsilon \sum_{j=1}^S {A}_{a,j}[\xi_j(t) - \xi_a(t)],
\end{align}
where $t$ is the number of update rounds, and $\epsilon$ is a given parameter controlling convergence. All the $S$ nodes perform this update by communicating with their neighbors.  In the matrix form, Eq.~\eqref{eq:Dynamics_elementwise} is written as
\begin{align}\label{eq:Dynamics_matrix}
\bm{\xi}(t+1) =\sfW_\epsilon \bm{\xi}(t) \;\;\; \mbox{with} \;\;\;
\sfW_\epsilon \triangleq \sfI_S  - \epsilon( \sfD-\sfA),
\end{align}
where $\sfD\triangleq\mathrm{diag}(d_1,\ldots,d_S)$ is the degree matrix with $d_s$ being the degree of the $s$-th node, $\sfI_S$ is the $S$-dimensional identity matrix, and $\bm{\xi}(t) \triangleq (\xi_1(t),\ldots,\xi_S(t))^\top$. 

The key idea of multi-agent coordination is to associate a stationary solution of the Markov transition defined by Eq.~\eqref{eq:Dynamics_matrix} with the process of consensus building. As can be easily verified, $\bm{u}_1 = \frac{1}{\sqrt{S}}\bm{1}_S$ is an $\ell_2$-normalized eigenvector of $\sfW_\epsilon$ whose eigenvalue is $\lambda_1=1$. If this is non-degenerated and the other absolute eigenvalues are less than one, Eq.~\eqref{eq:Dynamics_matrix} will converge to the stationary solution $\bm{\xi}^*$
\begin{align}\label{eq:achievingConv}
\bm{\xi}^* = {\sfW_\epsilon }^\infty  \bm{\xi}(0)
\approx {\lambda_1}^\infty \bm{u}_1 \bm{u}_1^\top  \bm{\xi}(0)= \frac{1}{S}\bar{\xi}\bm{1}_S 
\end{align}
because only the largest eigenvalue survives in the spectral expansion after an infinite number of transitions~\cite{Strang1976}. This means that all of the agents end up having the same value of $\frac{1}{S}\bar{\xi}$, which is the aggregation we wanted. We call this approach the \textit{dynamical consensus algorithm}. 

One obvious limitation of this approach is that the connected peers can see the original value in the first iteration, which is a privacy breach. To address this issue, we propose two secret sharing approaches: One is Shamir's secret sharing~\cite{shamir1979share} and the other is random chunking we proposed originally in~\cite{ide2019efficient}.

\subsection{Shamir's secret sharing}\label{subsec:Shamir}

In Shamir's secret sharing scheme, each of the $S$ agents first generates random numbers $R^a_1, \ldots, R^a_{S-1}$ in a large enough integer domain to define an $(S-1)$-th order polynomial function. For the $a$-th agent, the polynomial is defined by
\begin{align}
{g}_a(n) = \xi_a + R^a_1  n +\cdots + R^a_{S-1}  n^{S-1}.\label{eq:Shamir-what-is-sent}
\end{align}
Using this polynomial, each agent locally computes $S$ values $\{ {g}_s(1), \ldots, {g}_s(S) \}$. Then, by repeating the dynamical consensus algorithm $S$ times (i.e.,~by setting $\xi_s(0) = g_s(n)$ for $n=1,\ldots, S$ in Eq.~\eqref{eq:Dynamics_elementwise}), the agents build a consensus on $S$ aggregated values $\{ \bar{g}(1), \ldots, \bar{g}(S) \}$, where $\bar{g}(l) \triangleq \sum_{s=1}^{S}g_s(l)$. Since $R^a_1$ etc. are random numbers, raw data $\{\xi_a\}$ will not be revealed to the peers in the communication process.

At this point, each agent has the $S$ input-output pairs $\{ (1,\bar{g}(1)),\ldots, (S,\bar{g}(S)) \}$ of the polynomial 
\begin{align}
    \bar{g}(n) = \bar{\xi} + \bar{R}_1  n +\cdots + \bar{R}_{S-1}  n^{S-1}
\end{align}
at hand, where $\bar{R}_i \triangleq \sum_{a=1}^S  R^a_{i}$. Notice that we have $\bar{\xi}$ on the r.h.s.~as the intercept. Since an $(S-1)$-th order polynomial is uniquely determined by distinctive $S$ points, each agent can uniquely identify the functional form of $\bar{g}(n)$. With Lagrange's interpolation formula, the agents obtain the intercept by
\begin{align}\label{eq:LagrangeInterpolation}
\bar{\xi} = \bar{g}(0) = 
\sum_{l=1}^S \bar{g}(l) \prod_{m\neq l} \frac{m}{(m-l)}.
\end{align}

Shamir's algorithm is secure in the sense that it satisfies rigorous cryptographic conditions~\cite{boneh2016graduate}. However, when $S$ is large, the product form of Eq.~\eqref{eq:LagrangeInterpolation} needs a special attention to avoid numerical issues. Also, the need for $S$ repeated consensus makes it computationally less efficient, as will be empirically shown in Section~\ref{sec:Experiments}.

\subsection{Random chunking}\label{subsec:chunking}

Shamir's algorithm is based on the idea of recovering the data from seemingly non-informative multiple signatures. Here it is interesting to see what happens if we use a datum itself as the signature. Specifically, each of the agents secretly splits a local datum $\xi_s$ into a few chunks $\{ \xi_s^{[h]}\}$ such that $\sum_h \xi_s^{[h]} = \xi_s$. Then, the agents run the dynamical consensus algorithm on each of the chunks to get $\{\bar{\xi}^{[1]}, \ldots, \bar{\xi}^{[N_\mathrm{C}]}\}$, where $N_\mathrm{C} \ll S$ is the predefined number of chunks. Since aggregation is an linear operation, we have
\begin{align}
\bar{\xi}^{[1]}+\cdots+\bar{\xi}^{[N_\mathrm{C}]}
= \sum_{s=1}^S (\xi_s^{[1]}+\cdots +\xi_s^{[N^c]})
=\bar{\xi}.
\end{align}
This means that the total aggregation can be computed by performing chunk-wise aggregations without sharing the raw data $\{\xi_s\}$ explicitly. The procedure is summarized in Algorithm~\ref{algo:conChunk}. 

\begin{algorithm}[t]
\caption{Dynamical consensus with random chunking}
\label{algo:conChunk}
\begin{algorithmic}[1]
\STATE Input: $\epsilon$, $N_\mathrm{C}$. Initialize $\bar{\xi}=0$
\STATE Split $\xi_a$ into $N_\mathrm{C}$ chunks for $\forall a$.
\FOR {$i_\mathrm{C} \leftarrow 1, N_\mathrm{C}$}
\STATE Randomly generate $\sfA$. Confirm all the links are available for communication (re-generate $\sfA$ otherwise).
\STATE Initialize $\xi_a(0)$ as $\xi_a^{[i_\mathrm{C}]}$ for $\forall a$.
\REPEAT
\STATE Each agent synchronously perform~\eqref{eq:Dynamics_elementwise}.
\UNTIL{convergence}
\STATE $\bar{\xi} \leftarrow \bar{\xi} + \bar{\xi}^{[i_\mathrm{C}]}$
\ENDFOR
\end{algorithmic}
\end{algorithm}

Here, to prevent the agents from recovering the raw data by receiving all the $N_\mathrm{C}$ chunks, the chunk-wise aggregations must be made on different incidence matrices. There are two major solutions having different levels of intervention of the network infrastructure (see Section~\ref{subsec:setting}). For the higher intervention side, upon starting aggregation, the router may randomly pick $\sfA$ from a set of prepopulated incidence matrices, in which any pair of the graphs do not share the same edge. If the router always picks an $\sfA$ that is different from the previous aggregation round, $N_\mathrm{C}=2$ suffices.

If we cannot expect too much from the infrastructure, the security guarantee becomes probabilistic. This is reminiscent of Bitcoin's probabilistic security guarantee, which makes it the \textit{probabilistic finality} protocol~\cite{xiao2020survey}. In the next section, we discuss the detail of probabilistic guarantee of the random chunking algorithm.

\section{Privacy Breach Analysis}\label{sec:security}

This section provides probabilistic guarantees of the random chunking algorithm under three major attack scenarios. 

\subsection{Independent agent scenario}

As suggested in Algorithm~\ref{algo:conChunk}, one practical scenario is for the router to randomly assign the node name whenever starting aggregation, keeping the graph structure itself fixed. To study the risk of privacy breach, consider the event that the $j$-th node is the breach node to $s$, meaning that the node $j$ is able to fully recover $\xi_s$ by summing over $N_\mathrm{C}$ chunks it received. This happens when the $j$-th node stays as the neighbor over the $N_\mathrm{C}$ rounds, regardless of the other nodes. Since the $j$-th node gets chosen in each of the $N_\mathrm{C}$ chunking round with a probability 
\begin{align}
\binom{S-2}{d_s-1}\binom{S-1}{d_s}^{-1} =\frac{d_s}{S-1},
\end{align}
the probability that the $j$-th node is the breach node to $s$ regardless of the other nodes is given by $\left(\frac{d_s}{S-1}\right)^{N_\mathrm{C}}$. Since some of the other $S-2$ nodes may be the breach node as well, we have
\begin{align}\label{eq:P_breach-byChance}
p_\mathrm{breach}^s \leq \sum_{j \neq s}\left( \frac{d_s}{S-1} \right)^{N_\mathrm{C}} 
= (S-1)\left( \frac{d_s}{S-1} \right)^{N_\mathrm{C}}
\end{align}
by Boole's inequality (or the union bound). Therefore, we conclude that the probability of not having any breach node in the network is lower-bounded as
\begin{align}
p_\mathrm{secure} &\geq
\prod_{s=1}^S\left\{1- (S-1) \left( \frac{d_s}{S-1} \right)^{N_\mathrm{C}}\right\}\nonumber\\
&\geq 1- S(S-1) \left( \frac{d_\mathrm{max}}{S-1} \right)^{N_\mathrm{C}},\label{eq:p_secure_bound3}
\end{align}
where the second inequality is due to Bernoulli's inequality. The second term of the bound can be made arbitrarily small for \textit{sparse graphs} $d_\mathrm{max} \ll S$ by appropriately choosing the value of $N_\mathrm{C}$. To summarize, we have proved the following theorem: 
\begin{theorem}\label{prop:breach}
The dynamical consensus algorithm with random chunking has a privacy guarantee of Eq.~\eqref{eq:p_secure_bound3}. The probability of privacy breach can be made arbitrarily small. 
\end{theorem}

\subsection{Colluded agent scenario}\label{app:collusion_prob}

Another interesting question is whether the algorithm is secure under collusion. Although the risk of collusion is minimal in consortium-based networks, where the identity and the motivation of the agents are known to each other, we have the following guarantee:
\begin{theorem}\label{prop:collusion}
Let $N_\mathrm{L}$ be the number of colluded agents. The probability that the privacy of a non-colluded node (indexed by $s$) is compromised due to collusion is given by
\begin{align}
\label{eq:collusion_equality}
p_\mathrm{c.breach}^s &= \left\{
1 - \prod_{l=1}^{N_\mathrm{L}}\left(1 - \frac{d_s}{S-l}\right) \right\}^{N_\mathrm{C}}\\
&\leq \label{eq:collusion}
\exp\left\{-N_\mathrm{C} \left( 1 - \frac{d_s}{S-N_\mathrm{L}}\right)^{N_\mathrm{L}}\right\}
\end{align}
for $N_\mathrm{L} < S-d_s$, and 1 otherwise. 
\end{theorem}
\noindent
(\textit{Proof}) For an $s$-th non-colluded node, privacy breach occurs when at least one colluded node receives the chunk in all the $N_\mathrm{C}$ chunking rounds. Therefore, we have 
\begin{align}\label{eq:pL1}
    p_\mathrm{c.breach}^s = (1 -p_\mathrm{L})^{N_\mathrm{C}},
\end{align} 
where $p_\mathrm{L}$ is the probability that no colluded nodes are chosen in the neighbor set. The probability $p_\mathrm{L}$ can be evaluated as 
\begin{align}
p_\mathrm{L} &= \binom{S-1-N_\mathrm{L}}{d_s}\binom{S-1}{d_s}^{-1} \nonumber\\
&=
\frac{
(S-d_s-1)(S-d_s-2)\cdots (S-d_s-N_\mathrm{L})
}{
(S-1)(S-2) \cdots (S-N_\mathrm{L})
},\label{eq:pL2}
\end{align}
which holds only for $N_\mathrm{L} \leq S-d_s -1$. If there are so many colluded nodes that $N_\mathrm{L} \geq S-d_s$ holds, it is impossible not to choose a colluded nodes $N_\mathrm{C}$ times in a row and thus $p_\mathrm{L}$ must be zero. From Eqs.~\eqref{eq:pL1} and~\eqref{eq:pL2}, we have the equality of Theorem~\ref{prop:collusion}:
\begin{align}\label{eq:collusion-breach-equality}
p_\mathrm{c.breach}^s
&=  \left\{
1 - \prod_{l=1}^{N_\mathrm{L}} \left( 1 - \frac{d_s}{S-l}
\right)\right\}^{N_\mathrm{C}}.
\end{align}

Next, let us derive the upper bound. By replacing the product in Eq.~\eqref{eq:collusion-breach-equality} with the power of the minimum term and apply the Bernoulli inequality, we have
\begin{align}
p_\mathrm{c.breach}^s &\le \left\{
1 - \left( 1 - \frac{d_s}{S- N_\mathrm{L}} \right)^{N_\mathrm{L}}\right\}^{N_\mathrm{C}}\nonumber\\
&\leq
\exp\left\{-N_\mathrm{C} \left( 1 - \frac{d_s}{S-N_\mathrm{L}}\right)^{N_\mathrm{L}}\right\}.\label{eq:c-breach-better},
\end{align}
which completes the proof. $\square$

\begin{figure}[tb]
\begin{center}
\includegraphics[trim={0cm 0cm 0cm 0cm},clip,height=4.5cm]{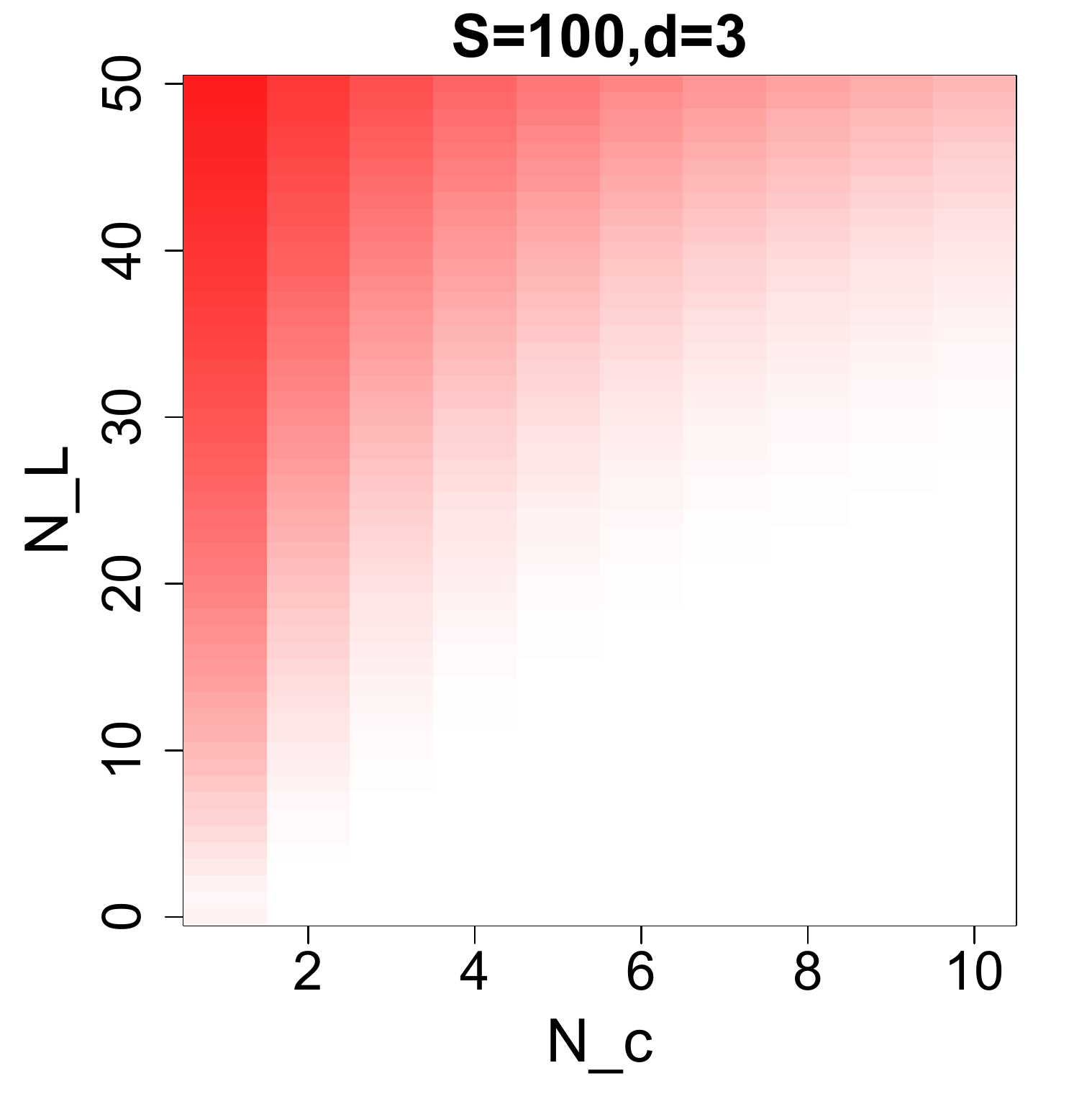}
\includegraphics[={0cm 0cm 0cm 0cm},clip,width=4.2cm]{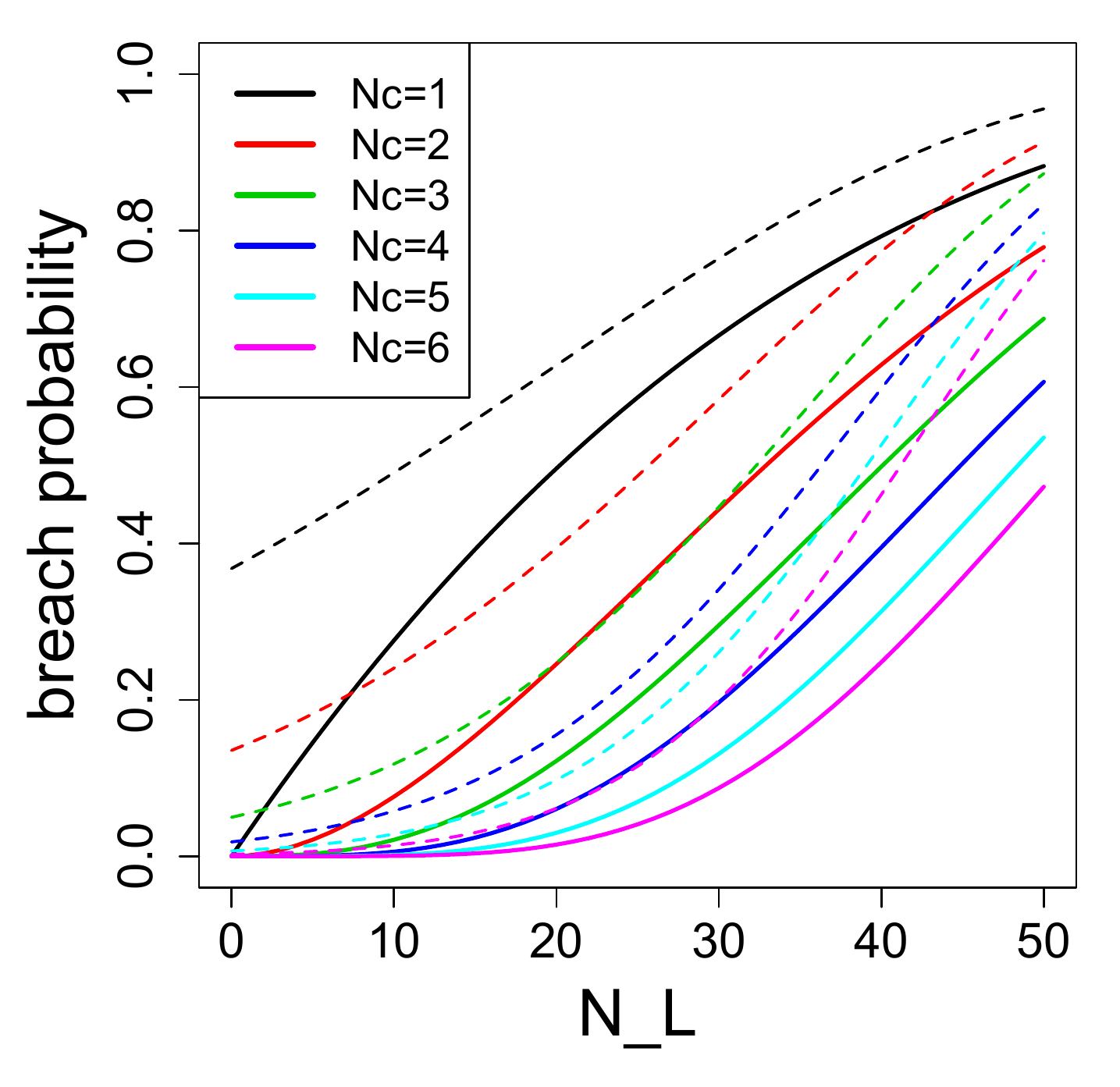}
\end{center}\vspace{-0.4cm}
\caption{Privacy breach probability due to collusion computed for $3$-regular graph of $S=100$. (a) Left panel: $p_\mathrm{c.breach}^s$ computed by Eq.~\eqref{eq:collusion_equality} with a gradation from red being 1 to white being 0. (b) Right panel: $p_\mathrm{c.breach}^s$ (solid lines) and its upper bound (Eq.~\eqref{eq:c-breach-better}; dashed lines) at a few $N_\mathrm{C}$ values. Best viewed in color.  }
\label{fig:CollusionBreach-bounds.pdf}
\end{figure}

Figure~\ref{fig:CollusionBreach-bounds.pdf}~(a) visualizes the privacy breach probability due to collusion (Eq.~\eqref{eq:collusion}) as a function of $N_\mathrm{C}$ and $N_\mathrm{L}$. As expected, the probability is almost one if $N_\mathrm{L}\sim \frac{S}{2}$ and $N_\mathrm{C} \sim 1$. However, we also see that the risk can be made negligible by properly choosing $N_\mathrm{C}$, as claimed in Theorem~\ref{prop:collusion}. Also, Fig.~\ref{fig:CollusionBreach-bounds.pdf}~(b) shows comparison among the exact probability and its upper bound computed by Eq.~\eqref{eq:c-breach-better}.

The inequality enables us to reasonably set the number of chunks, $N_\mathrm{C}$, so that the privacy of all non-colluded nodes is guaranteed with a sufficiently high probability, say, at least $1 - \eta$ for a small $\eta$. This corresponds to $p_\mathrm{c.breach}^s \leq \eta$ and is translated into
\begin{align}
N_\mathrm{C} \geq | \ln \eta | \left( 
1 - \frac{d_s}{S-N_\mathrm{L}}
\right)^{- N_\mathrm{L}}.
\end{align}
As long as the graph is sparse ($d_s \ll S$) and $N_\mathrm{L}$ is expected to be much smaller than $S$, a useful rule-of-thumb will be $2| \ln \eta |$. In consortium-based networks, $N_\mathrm{L}$ should be much smaller than $S$. In that case, the use of graphs having $d_s \ll S$ is critical to guarantee data privacy under the risk of collusion. Again, this motivates us to consider a family of \textit{sparse graphs}, as discussed in the next section.

\subsection{Eavesdropping scenario}
\label{app:eavesdropping}

Eavesdropping is one of the major attacks in communication network~\cite{karlof2003secure}. Although the risk of eavesdropping is limited in consortium-based networks, we can compute the probability of privacy breach in a similar way to the incidental privacy breach. The Shamir-based algorithm is safe even under eavesdropping and collusion because each agent sends only obfuscated values as Eq.~\eqref{eq:Shamir-what-is-sent} to the connected agents. 

In the random chunking algorithm, however, we again have a probabilistic guarantee. Consider the scenario that an eavesdropper picks a set of the edges and captures all the contents of communications. Let $N_\mathrm{E}$ be the maximum number of edges the eavesdropper can tap. Let $p^s_\mathrm{e.breach}$ be the probability that the $s$-th agent incurs a privacy breach under eavesdropping. A breach occurs when the eavesdropper receives all of the $N_\mathrm{C}$ chunks. This happens when the eavesdropper successfully tapped any of the $d_s$ edges from the $s$-th agent in every chunking round. Thus
\begin{align}\nonumber
p^s_\mathrm{e.breach} =& 
[1-\mathrm{Pr}(\mbox{No edges from $s$ are eavesdropped})]^{N_\mathrm{C}},
\\ \nonumber
 =& [
 1-\mathrm{Pr}(\mbox{All the $N_\mathrm{E}$ edges are chosen} 
 \\ \nonumber
 &\quad \quad \quad \mbox{from the other $E-d_s$ edges})]^{N_\mathrm{C}},  
 \\ \nonumber
&= \left\{
1- \binom{E-d_s}{N_\mathrm{E}}\binom{E}{N_\mathrm{E}}^{-1}
\right\}^{N_\mathrm{C}}
\\ \label{eq:eavasdrop_breach_prob_equality}
&= \left\{ 
1 -\prod_{l=0}^{d_s -1} \left( 1 - \frac{N_\mathrm{E}}{E - l}\right)
\right\}^{N_\mathrm{C}}
\end{align}
where $\mathrm{Pr}(\cdot)$ denotes the probability of the event specified by the argument, and $E \triangleq \sum_{i=1}^S d^i$ is the total number of edges of the graph. Using the Bernoulli's inequality as in Eq.~\eqref{eq:c-breach-better}, we have
\begin{align}
p^s_\mathrm{e.breach} 
& \leq
\left\{
1- \left(1 - \frac{N_\mathrm{E}}{E-d_s+1} \right)^{d_s}
\right\}^{N_\mathrm{C}}\nonumber
\\ \label{eq:eavasdrop_prob_bound}
&\leq \exp\left\{ -N_\mathrm{C} 
\left(1 - \frac{N_\mathrm{E}}{E-d_s+1} \right)^{d_s}  
\right\}.
\end{align}

We summarize this result as bellow:

\begin{theorem}\label{th:evasdropping}
The probability of privacy breach due to eavesdropping in the dynamic consensus algorithm with random chunking is upper-bounded as Eq.~\eqref{eq:eavasdrop_prob_bound}.
\end{theorem}

Note that for $d$-regular graphs, $E=dS$. Since $S \gg d_s$ for sparse graphs, as long as the infected edge ratio $\frac{N_\mathrm{E}}{E}$ is expected to be much smaller than one, we can always make $p^s_\mathrm{e.breach}$ negligible by choosing a sufficiently large $N_\mathrm{C}$. To keep $p^s_\mathrm{e.breach}$ less than $\eta >0$, we need
\begin{align}
N_\mathrm{C} \geq |\ln \eta| \left(1 - \frac{N_\mathrm{E}}{E-d_s+1} \right)^{-d_s} .
\end{align}
For sparse graphs, unless a majority of edges are tapped, $N_\mathrm{C} \sim 2|\ln \eta|$ again is a useful rule-of-thumb. 

\begin{figure}[tb]
\begin{center}
\includegraphics[={0cm 0cm 0cm 0cm},clip,width=4.5cm]{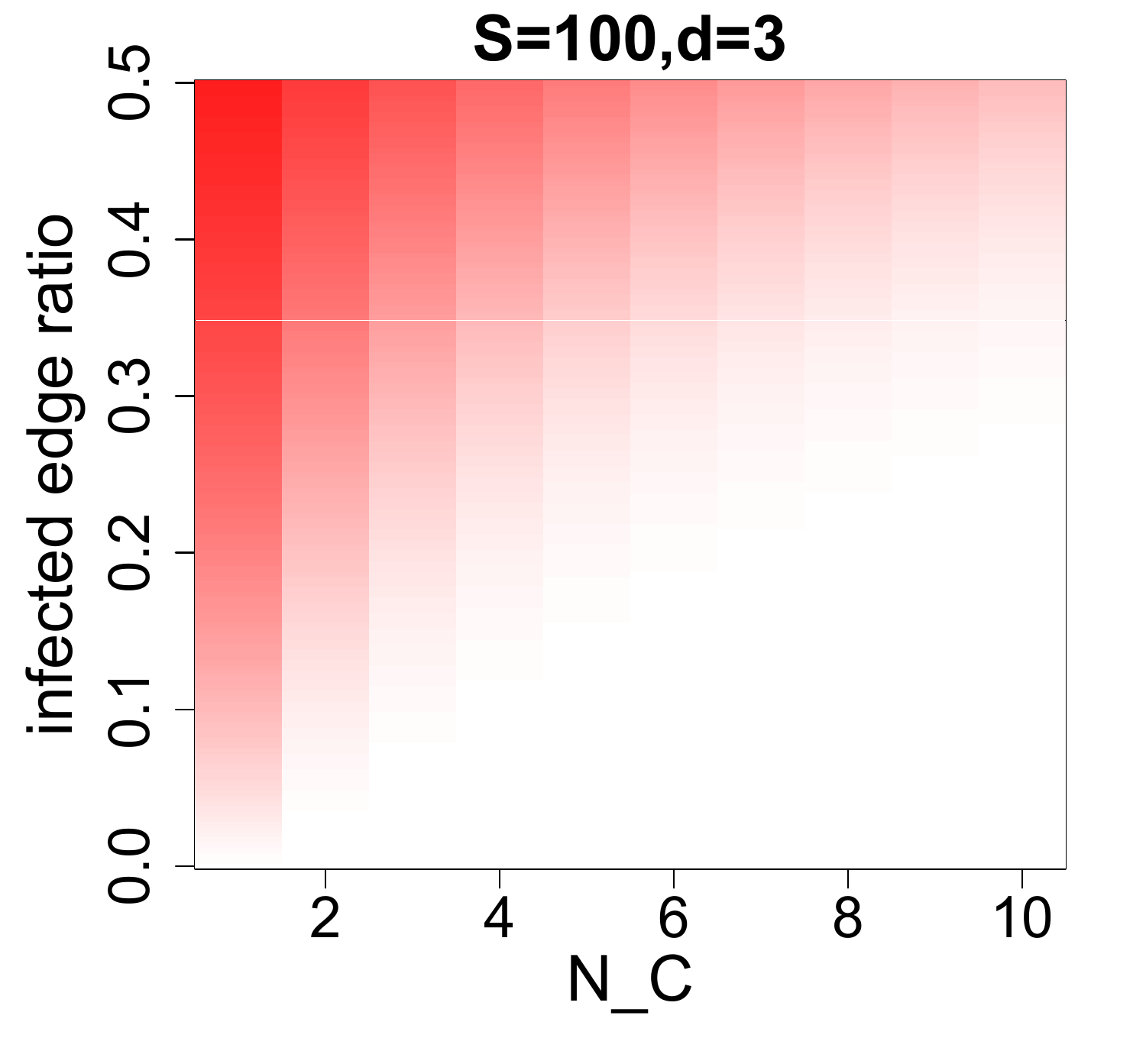}
\includegraphics[={0cm 0cm 0cm 0cm},clip,width=4.2cm]{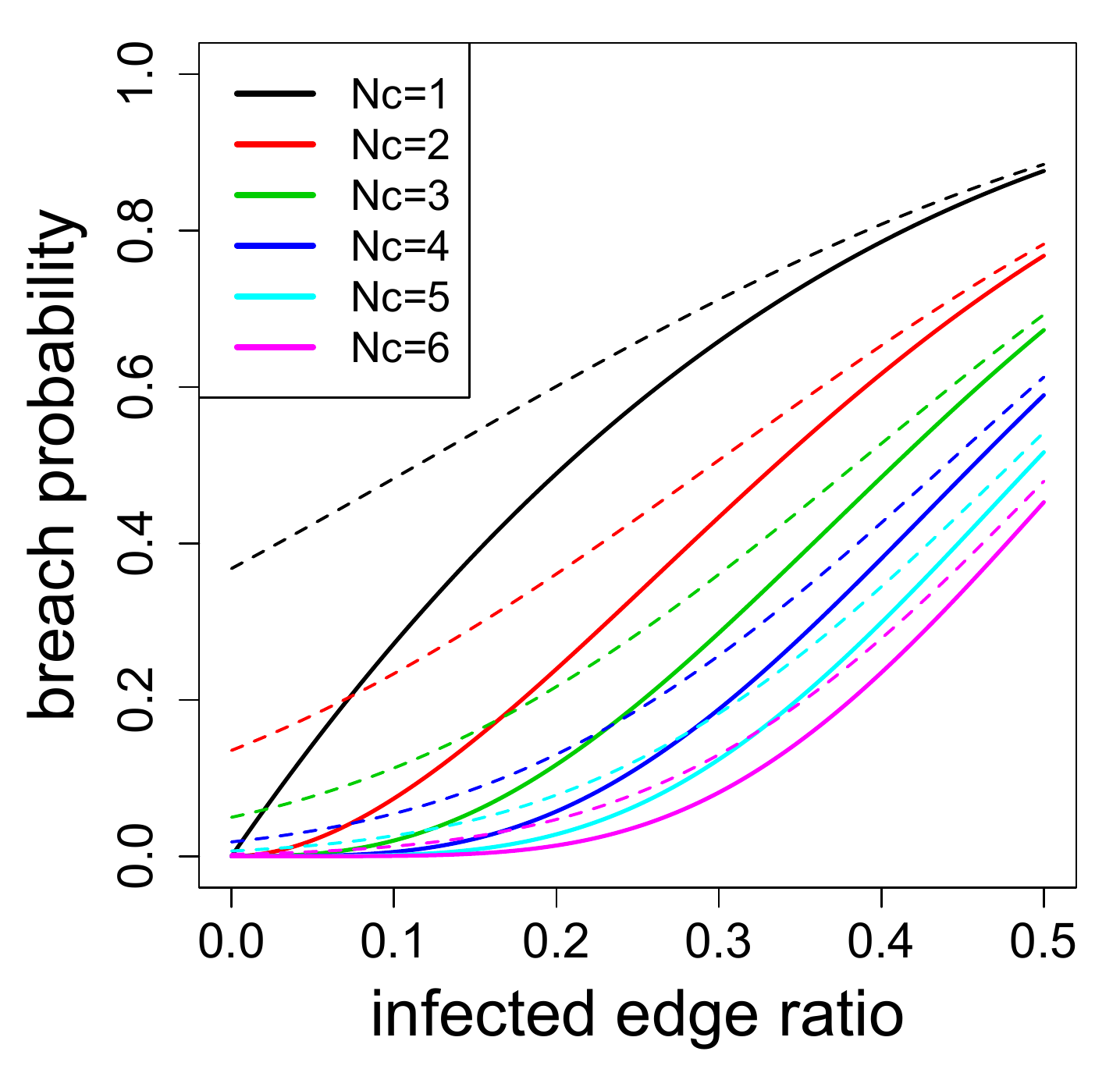}
\end{center}\vspace{-0.4cm}
\caption{Privacy breach probability due to eavesdropping computed for $3$-regular graph of $S=100$. (a) Left panel: $p^s_\mathrm{e.breach}$ computed by Eq.~\eqref{eq:eavasdrop_breach_prob_equality} as a function of $N_\mathrm{C}$ and the infected edge ratio $N_\mathrm{E}/E$ with a gradation from red being 1 to white being 0. (b) Right panel: $p^s_\mathrm{e.breach}$ (solid lines) and its upper bound (Eq.~\eqref{eq:eavasdrop_prob_bound}; dashed line) at a few values of $N_\mathrm{C}$. Best viewed in color.}
\label{fig:eaves-line}
\end{figure}

Figure~\ref{fig:eaves-line} illustrates $p^s_\mathrm{e.breach}$ as a function of $N_\mathrm{E}/E$ and $N_\mathrm{C}$ for a $3$-regular graph with $S=100$. As shown, even under massive eavesdropping where 20\% of the edges are tapped, $N_\mathrm{C}=6$ gives only about 1\% of breach risk.

In addition to the above case, we may consider the case where an eavesdropper hijacks an agent and capture all the outgoing communications. In this case, the random chunking algorithm is no longer secure although the Shamir-based method is still safe. This issue may be handled by combining with a more sophisticated secret sharing scheme such as in~\cite{gupta2017privacy}. Exploring this topic would be an interesting future topic.

\section{Network topology analysis}

In the previous section, we pointed out that the graph sparsity plays an important role in privacy preservation. This section discusses another important aspect of the network topology: \textit{Spectral} structure. Specifically, we discuss properties of a specific type of graph called the \textit{expander graph}. 

\subsection{Estimating the number of iterations to converge}

In the dynamic consensus algorithm discussed in Section~\ref{subsec:dynamicalConsensus}, the convergence speed is governed by the ratio between the first and second absolute largest eigenvalues. Let $\lambda_1=1, \lambda_2, \ldots, \lambda_S$ be the eigenvalues of $\sfW_\epsilon$ arranged in the decreasing order. For ease of exposition, let us assume that $\epsilon$ has been chosen so $\lambda_2$ is the absolute second largest as well, which is always possible. 

We are interested in the scalability of the algorithm in terms of the network size $S$. Let us first think about how the convergence rate is evaluated. Equation~\eqref{eq:achievingConv} shows that 
$$
\| \sqrt{S} { \sfW_\epsilon }^t  \bm{\xi}(0) \|_2 
\to \left\|  \frac{1}{\sqrt{S}}\bm{1}_S \bar{\xi} \right\|_2 = |\bar{\xi}|
$$
as $t \to \infty$, where $\|\cdot\|_2$ denotes the $\ell_2$ norm. Thus it makes sense to define the relative error at iteration round $t$ as
\begin{align}
e_\epsilon(t) \triangleq  \frac{1}{|\bar{\xi}|}\sqrt{\| \sqrt{S} { \sfW_\epsilon }^t  \bm{\xi}(0) \|_2^2  - \bar{\xi}^2 }
\end{align}
Let $\bmv_2$ be the $\ell_2$-normalized eigenvector corresponding to $\lambda_2$. As $t$ grows to infinity, we asymptotically have:
\begin{align}
e_\epsilon(t) \to  \sqrt{S} (\lambda_{2})^t \times \frac{\bmv_2^\top \bm{\xi}(0)}{\bar{\xi}}
\sim  \sqrt{S}(\lambda_{2})^t \times  \mathrm{O}(1).
\end{align}
Thus, we conclude that $\sqrt{S}(\lambda_{2})^t$ is a reasonable nondimensional metric for convergence. To achieve a relative error $\delta$, we need a number of iterations as
$
t \sim \mathrm{O}\left( \frac{\ln (\sqrt{S}/\delta)}{ |\ln\lambda_{2}|}    \right)
$.  
Let us summarize this result in Theorem~\ref{prop:iterationOrder}: 
\begin{theorem}\label{prop:iterationOrder}
In the dynamical consensus algorithm, the number of iterations $t$ to achieve a relative error $\delta$ is given by
\begin{align}\label{eq:number_of_iterations_wrt_S2}
    t \sim \mathrm{O}\left( \frac{\ln (\sqrt{S}/\delta)}{ |\ln(1 - \Delta_\lambda)|}    \right),
\end{align}
where $\Delta_\lambda \triangleq \lambda_1 - \lambda_2$. 
\end{theorem}
In Eq.~\eqref{eq:number_of_iterations_wrt_S2}, we have used the fact that $\lambda_1 =1$ and thus $\lambda_2 = 1 -\Delta_\lambda $. In our previous work~\cite{ide2019efficient}, we showed that the cycle graph scales quadratically as $t \sim S^2 \ln S$. The question is whether we can improve this (relatively poor) scalability by using a different communication graph. The next subsection answers this question.

\subsection{Expander graph}\label{subsec:expander}

Intuitively, we expect that the more information the agents circulate, the faster convergence we will get. The complete graph is clearly an extreme case, where the agents share information most generously with the peers and thus data privacy is least protected. Let $\mu_1,\ldots,\mu_S$ be the eigenvalues of $\sfA$ in the decreasing order. In the complete graph, it is easy to verify that $\mu_1 = S-1$ and $\mu_2 = \cdots = \mu_S=-1$. Hence, we have $\Delta_\lambda = \epsilon(\mu_1-\mu_2) = \epsilon S$. The complete graph is a $d$-regular graph with $d=S-1$. By Theorem~\ref{prop:iterationOrder}, with a choice of $\epsilon \sim \mathrm{O}(1/d)$, the convergence speed measured by $t$ scales as $\ln S$, which is much better than the quadratic scalability of the cycle graph. 

Is there any \textit{sparse} graph that behaves like the complete graph when communicating with the other agents? This may sound like a ridiculous question, but surprisingly, there exists a class of sparse graph called the \textit{expander graph} that has the same logarithmic convergence. Formally, the expander graph is defined as a graph whose \textit{expansion constant} is lower-bounded, where 
\begin{align}
\mbox{(expansion constant)}\triangleq
   \inf_{\mathcal{V}_1}\left\{
   \frac{|\partial \mathcal{V}_1|}{\min\{|\mathcal{V}_1|, S-|\mathcal{V}_1|\}} \right\} 
\end{align}
with $\mathcal{V}_1$ being an arbitrary subset of the graph nodes and $|\partial \mathcal{V}_1|$ is the number of outgoing edges from the subgraph~\cite{davidoff2003elementary}. Intuitively, the more the expansion constant is, the more ``talkative'' the nodes tend to be. Very interestingly, this purely geometric definition has a hidden connection to the graph spectrum, and, in our case, we have
\begin{align}\label{eq:Delta_lambda_expander}
    \Delta_\lambda \geq \epsilon \frac{\alpha^2}{2d} \quad \mbox{($d$-regular expander graphs)}
\end{align}
 by Cheeger's inequality~\cite{brouwer2011spectra}, where $\alpha$ is the lower bound of the expansion constant. Thus we again have the logarithmic scalability $t\sim \ln S$ with a choice of $\epsilon \sim \mathrm{O}(1/d)$. We now summarize the result as follows. 
\begin{theorem}\label{prop:expander}
On $d$-regular expander graphs, $t \sim \mathrm{O}(\ln (\sqrt{S}/\delta))$ with a choice of $\epsilon \sim \mathrm{O}(1/d)$.
\end{theorem}

\begin{figure}[t]
\begin{center}
\includegraphics[trim={0cm 0cm 0cm 0cm},clip,height=2.5cm]{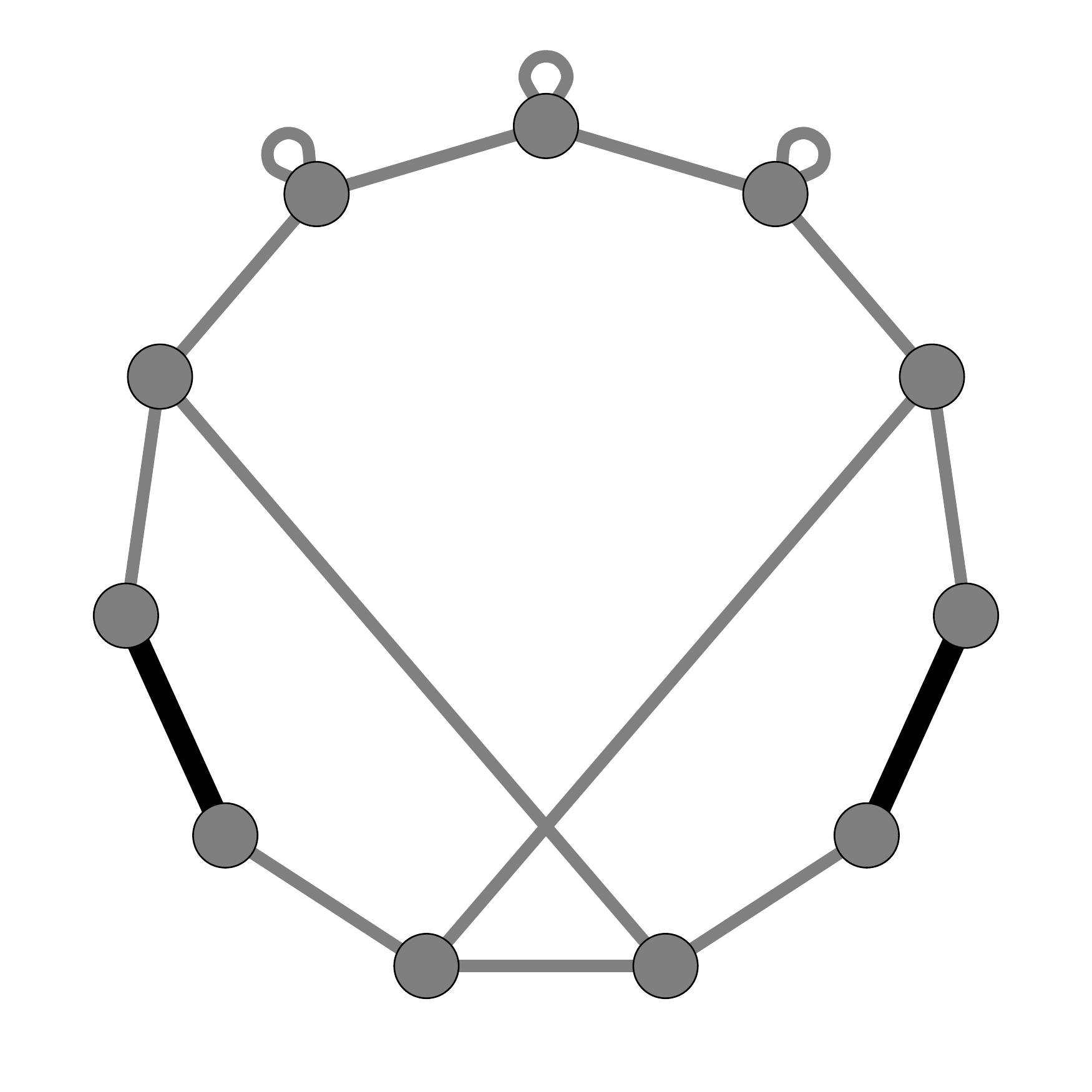}
\includegraphics[trim={0cm 0cm 0cm 0cm},clip,height=2.5cm]{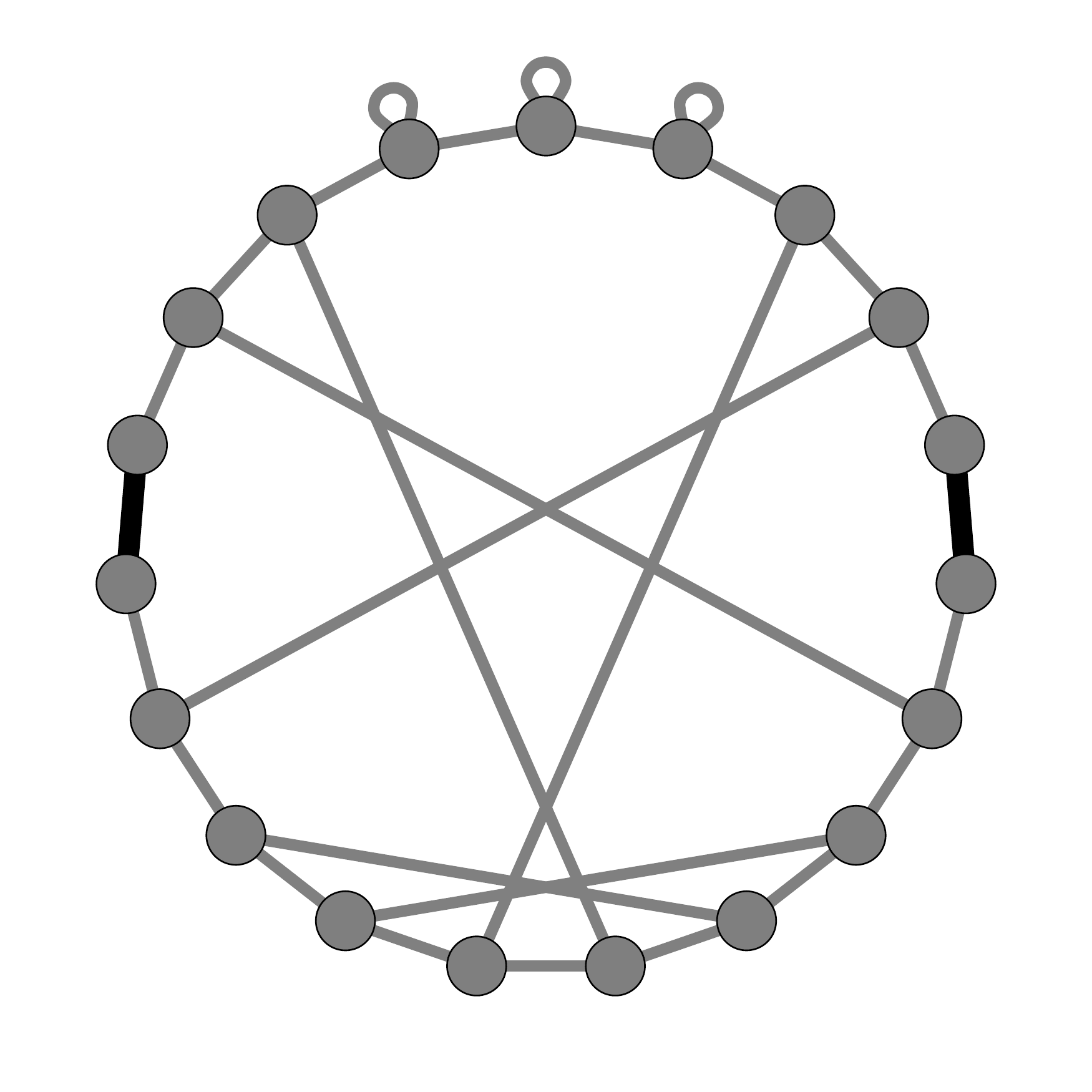}
\includegraphics[trim={0cm 0cm 0cm 0cm},clip,height=2.5cm]{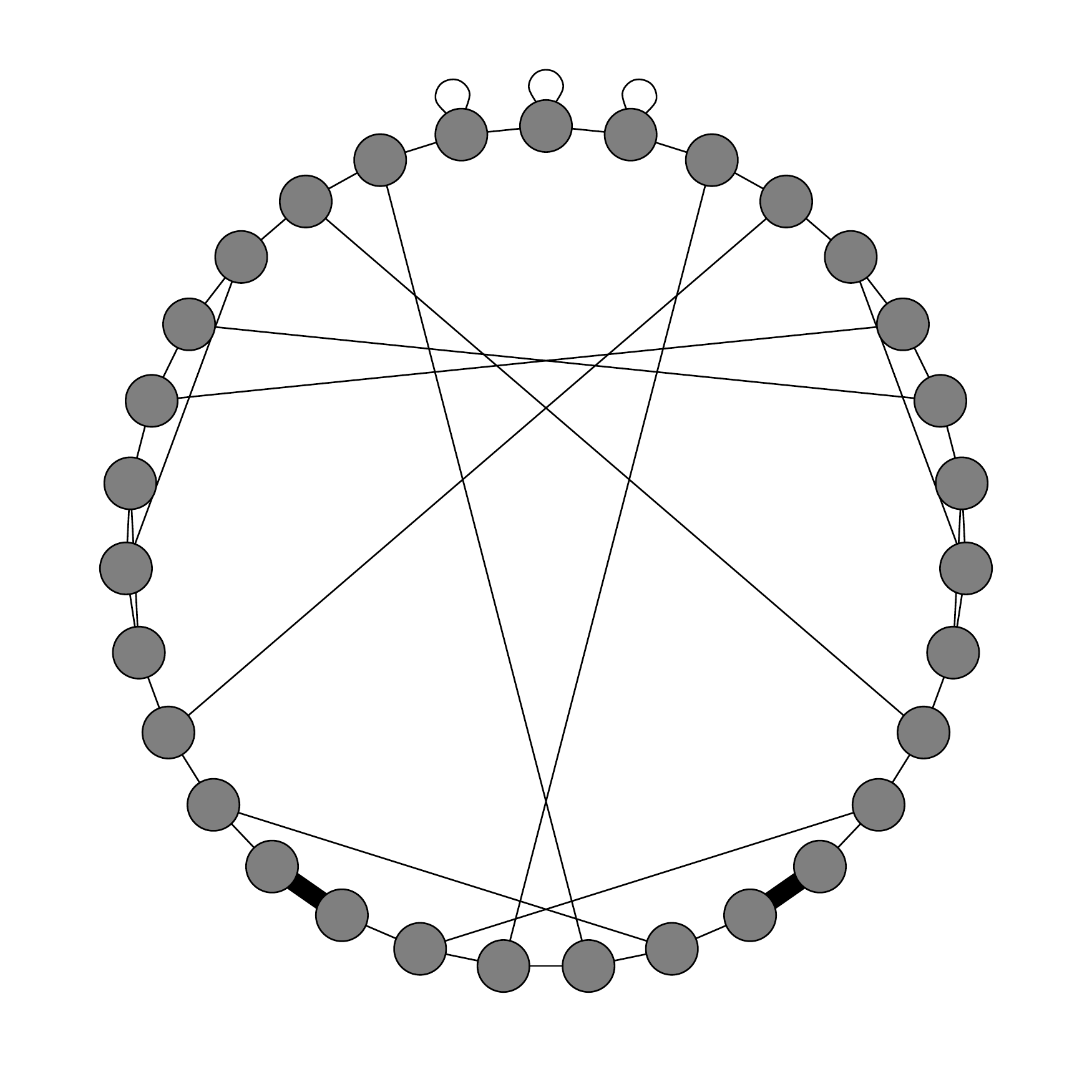}
\includegraphics[trim={0cm 0cm 0cm 0cm},clip,height=2.5cm]{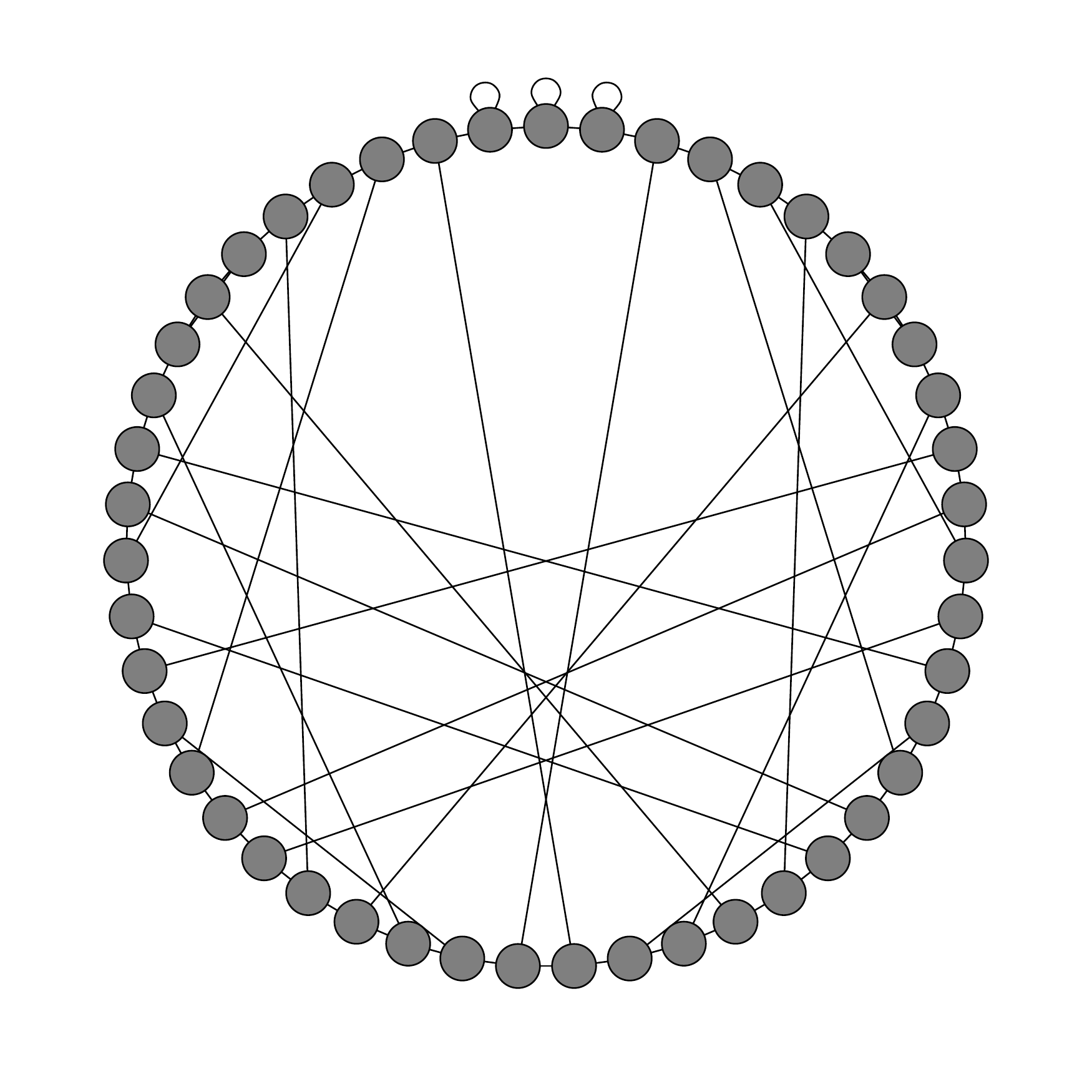}
\includegraphics[trim={0.75in 2in 0.75in 2in},clip,height=2.5cm]{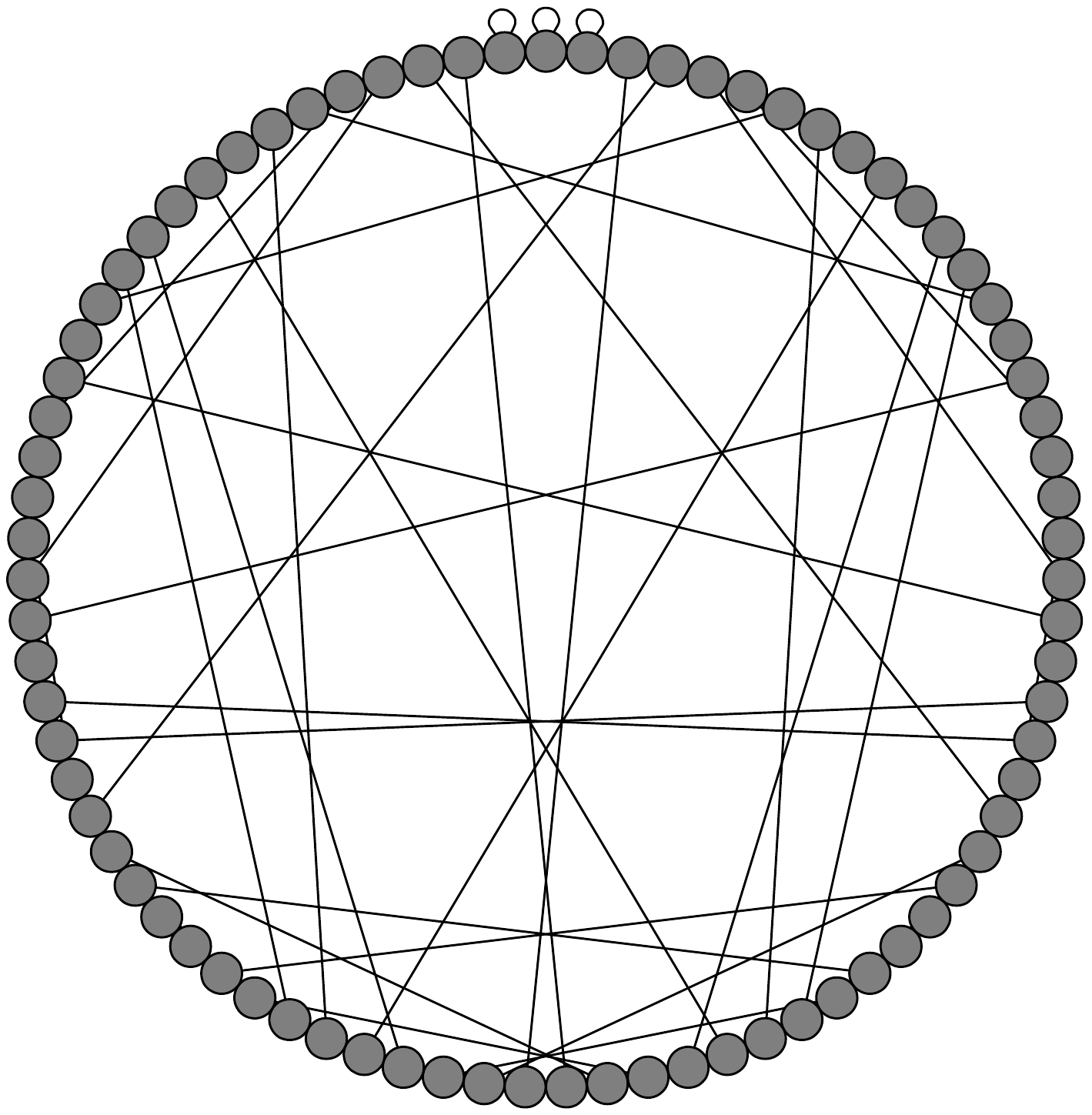}
\includegraphics[trim={0.75in 2in 0.75in 2in},clip,height=2.5cm]{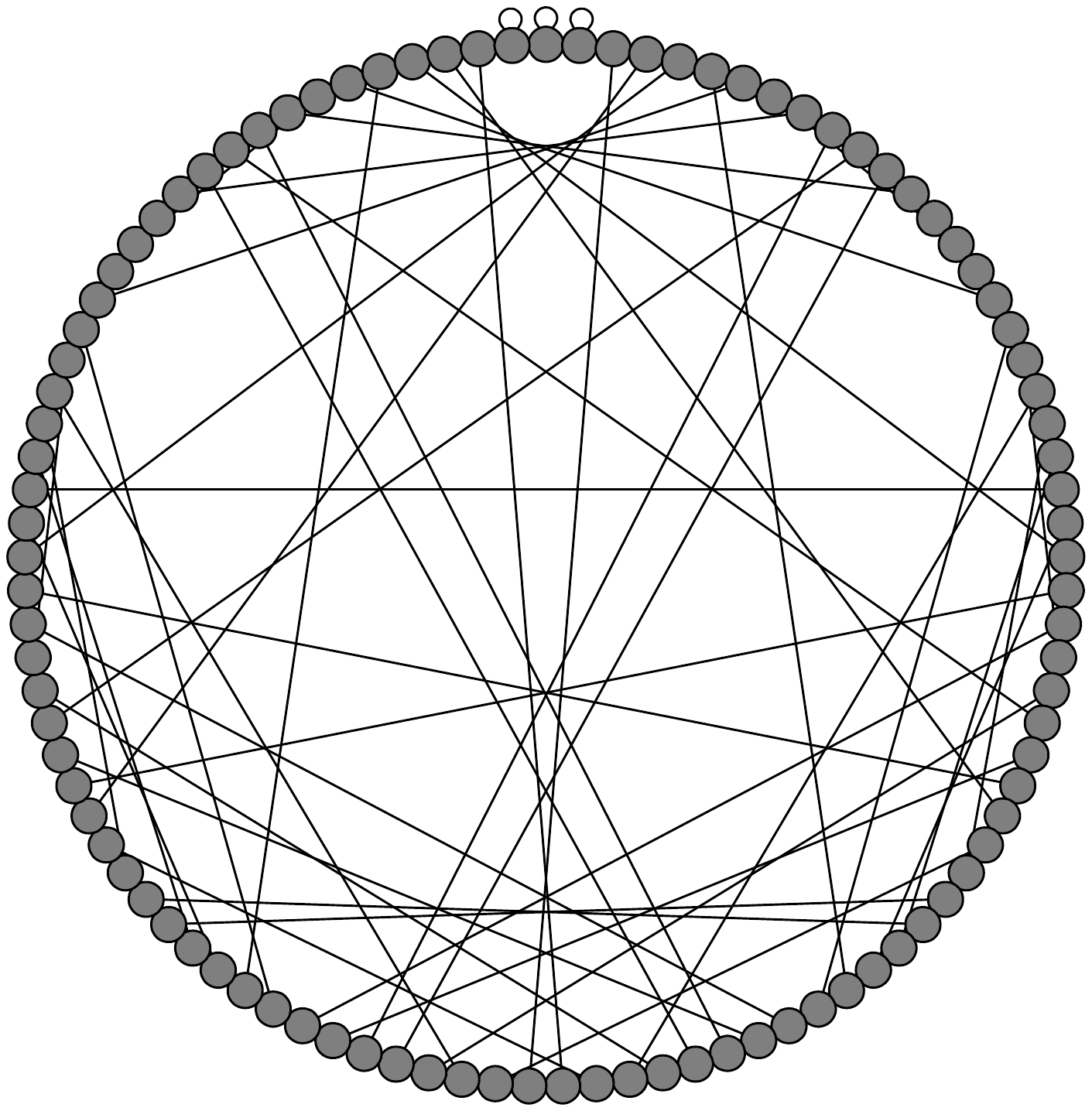}
\end{center}\vspace{-0.4cm}
\caption{Example of 3-regular expander graphs with $S=11,19,31$ (top row) and $47, 79, 97$ (bottom row) from left to right. Note that self-loops and double edges (thick lines) exist.}
\label{fig:ExpanderExample}
\end{figure}

Construction of expander graphs is nontrivial. Fortunately, there are two ways available to obtain expander graphs for our purpose. The first one is a $3$-regular expander graph called the cycle with inverse chords~\cite{vadhan2012pseudorandomness}. As the name suggests, it starts with the cycle graph (2-regular graph), and connects each node $s$ to the nodes $j = s\pm 1, (s-1)(j-1) =1 \mod S$ for a prime $S$. See Figs.~\ref{fig:S7expander_model.pdf} and~\ref{fig:ExpanderExample} for a few examples. As mentioned before, the node indices need to be randomly shuffled prior to each round of the random chunking algorithm. Since the eigenspectrum is invariant to re-labeling the nodes, it does not affect the convergence of dynamic consensus.

The other possible approach is to use a random graph. It is known that uniformly sampled $d$-regular random graphs approximate $\Delta_\lambda$ of the expander graph Eq.~\eqref{eq:Delta_lambda_expander} almost always perfectly~\cite{Friedman2004A}. To construct such a graph, we can leverage the configuration model in random graph theory~\cite{kim2006generating}. Specifically, each agent simply picks $d$ neighbors randomly and uniformly to form a $d$-regular graph. This approach is preferable in the sense that it does not require any intervention of the network router and does not assume network connection that is stable throughout the entire consensus process. On the other hand, however, the randomness in the network structure may lead to some unpredictability in the eigenspectrum, which is in contrast to the first option. To avoid extra randomness, we use the cycle with inverse chords in the empirical study in the next section. Evaluating and designing random graph construction algorithms in dynamic consensus is an interesting future research topic. 


\section{Experiments}\label{sec:Experiments}

This section reports on experimental results of the proposed model. All the experiments were conducted locally on a laptop PC with a Core i7 Processor and 32 GB memory. 

\subsection{Number of iterations to converge}

\begin{figure}[t]
\begin{center}
\includegraphics[trim={0cm 0cm 0cm 0.5cm},clip,height=6cm]{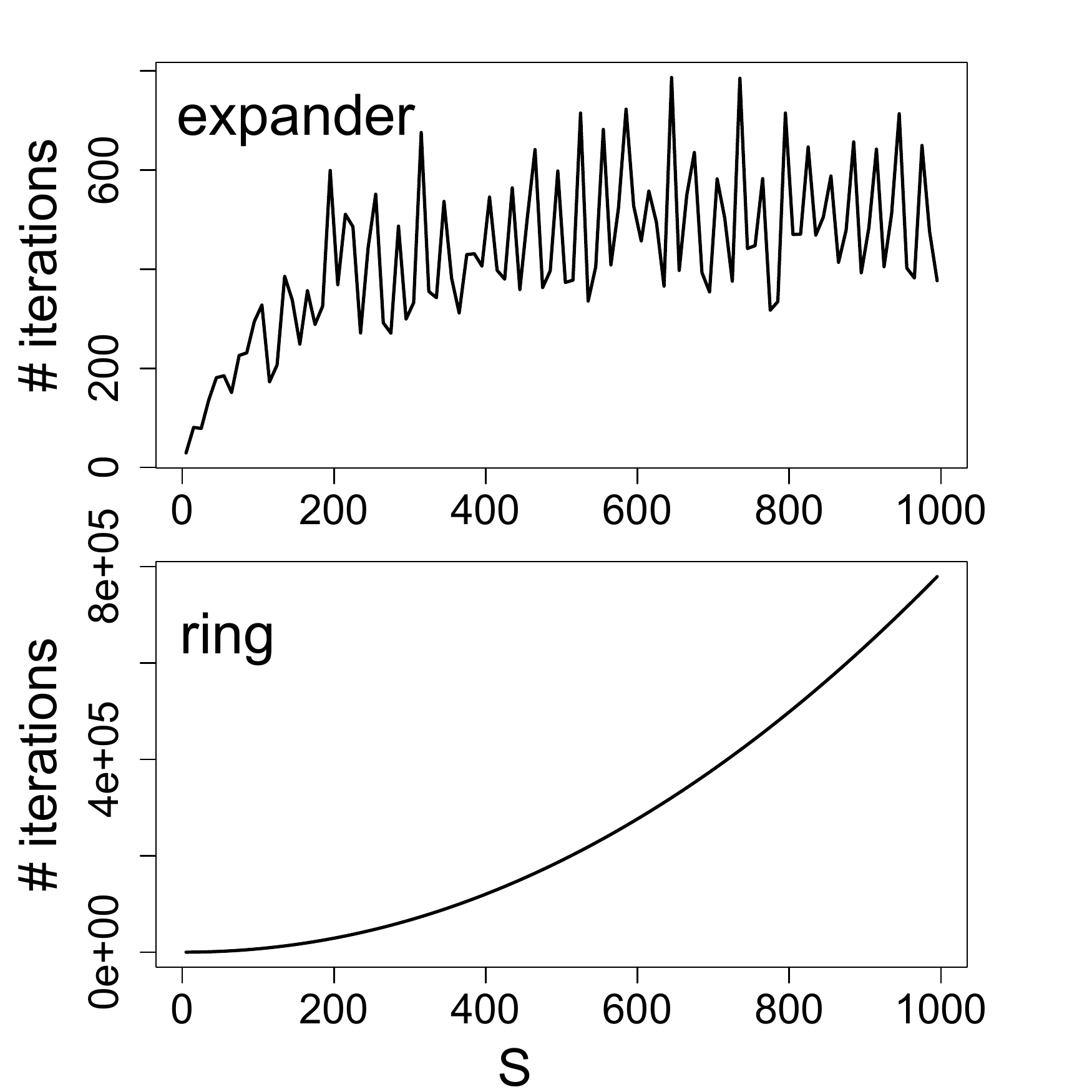}
\end{center}\vspace{-0.4cm}
\caption{Comparison of the number of iterations $t$ for $\delta = 10^{-3}$. }
\label{fig:experiments}
\end{figure}

Figure~\ref{fig:experiments} compares between the expander graph (cycle with a random chord) and the 2-regular cycle graph (or the ring) on the number of iterations $t$ to achieve $\delta = 10^{-3}$ as defined in Theorem~\ref{prop:iterationOrder} and~\ref{prop:expander}. It is interesting to observe that just adding an extra edge to the ring by the rule $(s-1)(j-1) =1 \mod S$ drastically changes the convergence behavior. As mentioned before, the ring scales quadratically as $t \sim S^2 \ln S$. In contrast, $t$ grows very slowly in the expander graph, being consistent to the logarithmic scalability predicted by Theorem~\ref{prop:expander}.

The original construction of the 3-regular expander graph is for $S$ that is a prime. One interesting question in practice is that the excellent convergence behavior is maintained for non-prime $S$'s. We extended the original construction by giving a self-loop whenever the equation $(s-1)(j-1)=1 \mod S$ does not have a solution, so that the graph is always $3$-regular. Fortunately, apart from the visible fluctuations, the figure suggests that the overall convergence behaviors are robust and we should be able to safely use the model even for non-primes.

\subsection{Shamir vs.~random chunking}\label{subsec:Shamir_vs_chunking}

Since one iteration needs one matrix-vector multiplication (see Eq.~\eqref{eq:Dynamics_matrix}), actual computational time may have a different scalability from that of the number of iterations to converge. Specifically, one may expect an extra $S\bar{d}$ factor, where $\bar{d}$ is the mean degree. 

\begin{figure}[tb]
\begin{center}
\includegraphics[trim={0cm 0cm 0cm 0cm},clip,height=4.28cm]{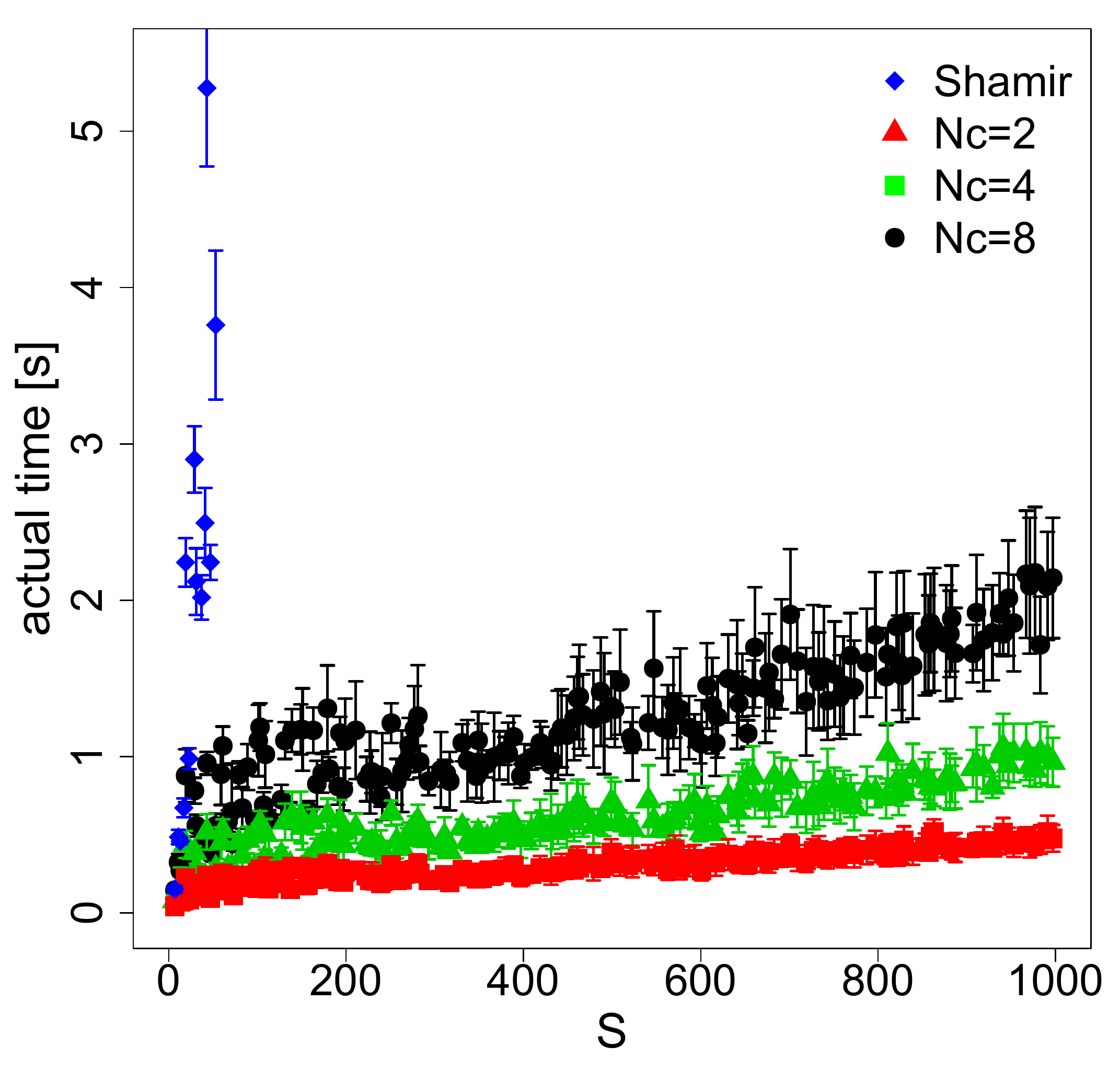}
\includegraphics[={0cm 0cm 0cm 0cm},clip,width=4.28cm]{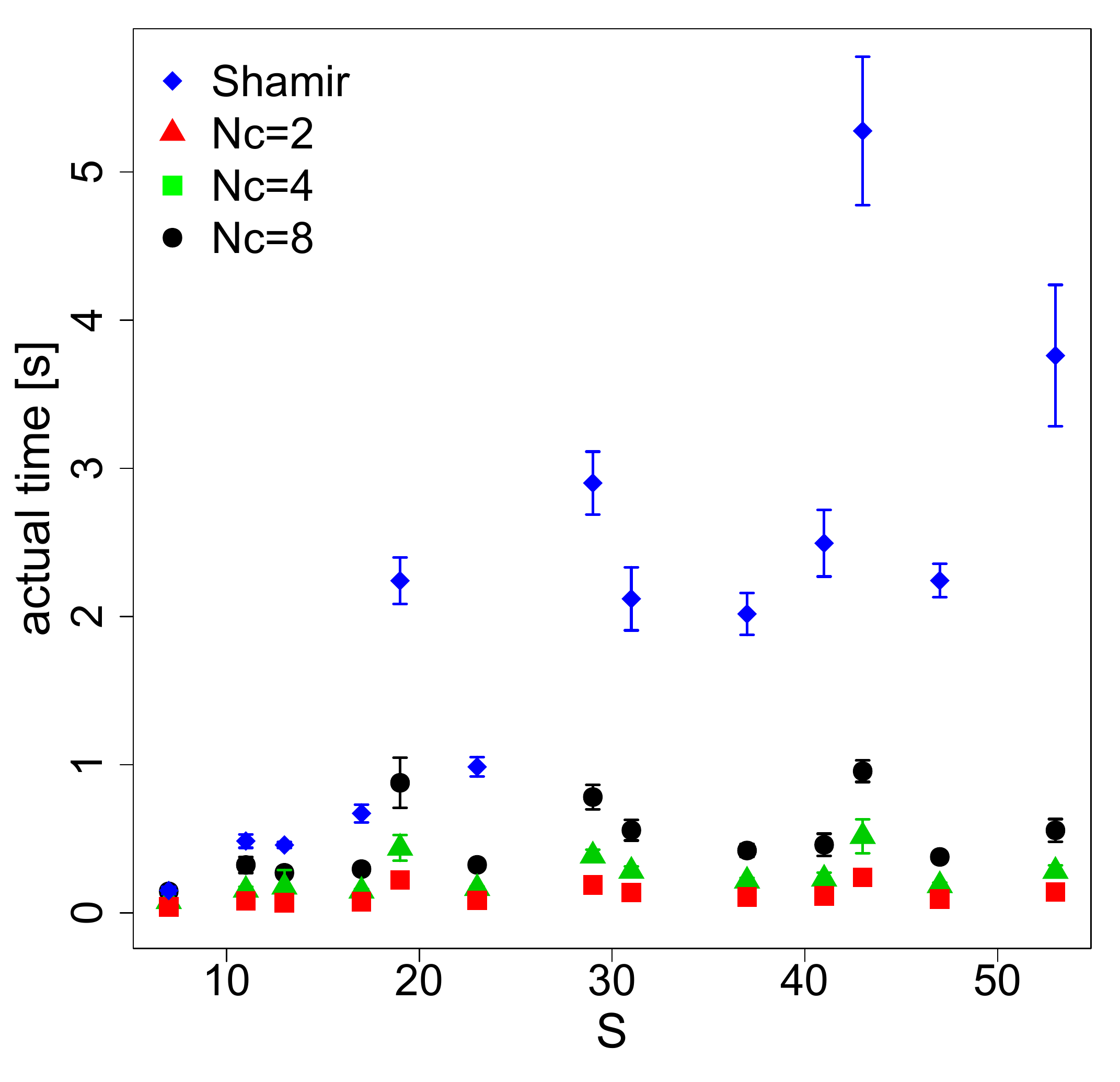}
\end{center}\vspace{-0.4cm}
\caption{Actual computation time for aggregation on the $3$-regular expander graph. The right panel covers the range of $7 \leq S \leq 53$. }
\label{fig:ActualTime_Shamir-chunk_S7to53.pdf}
\end{figure}

Figure~\ref{fig:ActualTime_Shamir-chunk_S7to53.pdf} compares actual computation times in different $N_\mathrm{C}$s. We randomly initialized $\xi_s(0)$ with the uniform distribution in $[-1,2]$. Convergence was declared when the relative error gets smaller than $10^{-5}$. The mean and standard deviation (s.d.) were computed by repeating computation by 10 times with a different random seed. The figure confirms the expected \textit{linear} dependency in the random chunking algorithm.

Figure~\ref{fig:ActualTime_Shamir-chunk_S7to53.pdf} also compares the Shamir's secret sharing algorithm with the random chunking algorithm. Since the Shamir-based method needs to repeat the process of global consensus $S$ times, the actual computation time is expected to depend quadratically on $S$, which is consistent with the figure. Apart from $S \lesssim 10$, where the both methods are comparable, the Shamir-based method tended to have a larger variability than the random chunking algorithm, which might suggest numerical issues in the implementation. 

Finally, we comment on the origin of non-smoothness of the computational time in Fig.~\ref{fig:ActualTime_Shamir-chunk_S7to53.pdf}. In the figure, close inspection shows that the computational times are correlated: whenever the Shamir-based method gets much time, so does the random chunking method. This suggests that the source of the non-smoothness is in the spectral structure of the underlying network because the same dynamical consensus algorithm was shared by both. Although we have obtained a very solid result on the convergence behavior, how the spectral gap $\Delta_\lambda$ scales as $S$ increases is not straightforwardly predictable. Further investigation on this point is left to future work.

\subsection{Shamir vs.~homomorphic encryption}

Table~\ref{table:BeatHE} compares computation time for a few small $S$'s between the proposed Shamir-based method and homomorphic encryption (HE)-based method of~\cite{ruan2017secure}. Both approaches have cryptographic security and share the same dynamical consensus algorithm. For a fair comparison, we ran both on the same expander graph and used the same experimental design as above. For HE, we used an implementation of the Paillier cryptosystem~\cite{homomorpheR}, which encrypts and decrypts every communication between the agents with a new key. From the table, and in the light of Fig.~\ref{fig:ActualTime_Shamir-chunk_S7to53.pdf}, we conclude that the proposed random chunking-based method is several orders of magnitude faster than the HE-based alternative. 

\begin{table}[bt]
\caption{Comparison of actual computation times in aggregation [sec].}\label{table:BeatHE}
\vspace{-1pt}
\centering
\begin{tabular}{c|cccc}
\toprule
 & \multicolumn{2}{c}{Shamir}  & \multicolumn{2}{c}{HE} \\ \cline{2-5}
$S$ 	& mean	& s.d.	& mean	& s.d. \\ \midrule
7 &  0.159 & 0.024 & 155	& 18.8 \\ 
11 &  0.485 & 0.045 & 288	& 44.1 \\ 
13 &  0.458 & 0.020 & 375	& 24.8 \\ 
17 &  0.671 & 0.060 & 574	& 55.1 \\ 
19 & 2.24 & 0.156 & 699	& 47.0 \\  
\bottomrule
\end{tabular}
\vspace{-1pt}
\end{table}

\subsection{Remarks on synchronization issues}
\label{app:smallS-Shamir-RC}

In Section~\ref{subsec:Shamir_vs_chunking}, we mentioned that actual computational time of the random chunking algorithm on the expander graph should have a linear dependency on $S$. This statement implicitly assumes that computation is made sequentially. Although in theory it is true that the agents can perform updates in parallel, sequential execution should be a reasonable assumption for evaluating computational time. In real distributed environments, we always need to handle the issue of synchronization across the network. If the cost of local computation is negligible, network delays will almost always dominate the time required to move on to the next iteration. In that case, most of the agents would spend most of the time waiting for the last message to be delivered to one of the agents. 

On the other hand, if the computational cost is on average much higher than network delays, we will need to consider parallelization as a realistic option. This is actually the case in homomorphic encryption (HE)-based secure computation, as recently pointed out in~\cite{liu2020secure}. There may also be some room for improvement in the HE implementation itself, where we employed~\texttt{homomorpheR}~\cite{homomorpheR} with the key size of $1\,024$ to follow the protocol proposed in~\cite{ruan2017secure,RuanGaoWang2019}, although handling signed floating-point numbers can be a subtle issue (see, e.g.~\cite{cheon2017homomorphic}). The key exchange protocol can be simplified, too. Regarding Table~\ref{table:BeatHE}, it would be an interesting future research topic to compare the random chunking method with a highly optimized HE implementation in a realistic setting.

\section{Concluding remarks}\label{sec:Conclusion}

We have presented new research directions of Blockchain as a collaborative value co-creation platform rather than a mere immutable data storage. Our platform is designed to respect the values of democracy, diversity, and privacy in its collaborative learning process. As such an instance, we have proposed a multi-task federated learning framework combined with a global consensus-building algorithm.

We discussed two topics in the global consensus-building process. The first topic was data privacy. We proposed two secure consensus-building approaches built upon the idea of secret sharing: The Shamir-based dynamic consensus, which has a cryptographic security guarantee, and the random chunking algorithm, which falls into the category of the probabilistic finality protocol. For the latter, we have provided the upper bound of privacy breach under three different attack scenarios. The second topic discussed was the issue of network design in global consensus. Based on the profound results in spectral graph theory, we showed that expander graphs dramatically improve the scalability of the algorithm.

We conclude this paper by summarizing a few future research topics:

\paragraph{Learning under network errors}
We have assumed perfectly synchronized and stable communication among the agents. As suggested in Sections~\ref{subsec:expander} and~\ref{app:smallS-Shamir-RC}, extensions to include network errors is of primary importance to make our platform truly useful.

\paragraph{Meta-agreement issues}
Another potential issue in practice is how to agree on the learning task itself (choice of the algorithm, data dimensionality $M$, the definition of each dimension, etc.) and how to initiate peer-to-peer communication (who to communicate with).

\paragraph{External data privacy}
One interesting business scenario is to sell the learned model to external parties. This calls for a guarantee that is different from the \textit{internal} data privacy among the agents. It is known that the classical concept of differential privacy has issues in evaluating noisy real-valued variables~\cite{minami2016differential}. Establishing a method of evaluating the degree of information leak is important. 

\paragraph{Randomness in graph spectra}
As discussed in Section~\ref{subsec:expander}, the origin of the non-smoothness shown in Fig.~\ref{fig:ActualTime_Shamir-chunk_S7to53.pdf} and the consensus algorithm based on random graphs are still an open question. 

\paragraph{Security analysis}
The random chunking algorithm combined with the dynamic consensus  algorithm appears to have more flexibility than traditional cryptographic methods. We need to study further the pros and cons of those methods. 

\paragraph{Use-cases} 
Finally, we need to develop practical use-cases where the decentralized architecture is truly useful. The lightweight probabilistic privacy guarantee seems suitable in internet-of-things (IoT) applications~\cite{ide2018collaborative}, but more study is needed. We hope that, just as Nakamoto's Bitcoin changed the landscape of financial transaction management, our decentralized collaborative learning platform as the next-generation Blockchain has the potential to change the way of doing business. 

\begin{small}
\section*{Acknowledgement}
T.I.~thank Dr.~Sachiko Yoshihama for her insightful suggestions. 
T.I.~is partially supported by the Department of Energy National Energy Technology Laboratory under Award Number DE-OE0000911. A part of this report was prepared as an account of work sponsored by an agency of the United States Government. Neither the United States Government nor any agency thereof, nor any of their employees, makes any warranty, express or implied, or assumes any legal liability or responsibility for the accuracy, completeness, or usefulness of any information, apparatus, product, or process disclosed, or represents that its use would not infringe privately owned rights. Reference herein to any specific commercial product, process, or service by trade name, trademark, manufacturer, or otherwise does not necessarily constitute or imply its endorsement, recommendation, or favoring by the United States Government or any agency thereof. The views and opinions of authors expressed herein do not necessarily state or reflect those of the United States Government or any agency thereof.
\end{small}

\bibliographystyle{IEEEtran}
\bibliography{ide_et_al}

\end{document}

%% file: ide_preamble.tex
\usepackage{algorithm}
\usepackage{algorithmic}
\usepackage{wrapfig}
\usepackage{graphicx}
\usepackage{mathtools} 
\usepackage{amssymb}
\usepackage{amsmath}
\usepackage{amsthm} 
\usepackage{color}
\usepackage{bm}

\newtheorem{theorem}{Theorem}

\usepackage{tcolorbox}

\newcommand{\bmu}{\bm{u}}
\newcommand{\bmv}{\bm{v}}

\newcommand{\bmx}{\bm{x}}

\newcommand{\bmz}{\bm{z}}

\newcommand{\bmT}{\bm{T}}
\newcommand{\bmU}{\bm{U}}

\newcommand{\bmtheta}{\bm{\theta}}

\newcommand{\bmmu}{\bm{\mu}}

\newcommand{\bmpi}{\bm{\pi}}

\newcommand{\sfA}{\mathsf{A}}

\newcommand{\sfD}{\mathsf{D}}

\newcommand{\sfI}{\mathsf{I}}

\newcommand{\sfW}{\mathsf{W}}

\newcommand{\calO}{\mathcal{O}}

%% file: ide_et_al.bbl
\begin{thebibliography}{10}
\providecommand{\url}[1]{#1}
\csname url@samestyle\endcsname
\providecommand{\newblock}{\relax}
\providecommand{\bibinfo}[2]{#2}
\providecommand{\BIBentrySTDinterwordspacing}{\spaceskip=0pt\relax}
\providecommand{\BIBentryALTinterwordstretchfactor}{4}
\providecommand{\BIBentryALTinterwordspacing}{\spaceskip=\fontdimen2\font plus
\BIBentryALTinterwordstretchfactor\fontdimen3\font minus
  \fontdimen4\font\relax}
\providecommand{\BIBforeignlanguage}[2]{{%
\expandafter\ifx\csname l@#1\endcsname\relax
\typeout{** WARNING: IEEEtran.bst: No hyphenation pattern has been}%
\typeout{** loaded for the language `#1'. Using the pattern for}%
\typeout{** the default language instead.}%
\else
\language=\csname l@#1\endcsname
\fi
#2}}
\providecommand{\BIBdecl}{\relax}
\BIBdecl

\bibitem{tse2017blockchain}
D.~Tse, B.~Zhang, Y.~Yang, C.~Cheng, and H.~Mu, ``Blockchain application in
  food supply information security,'' in \emph{Proceeding of 2017 IEEE
  International Conference on Industrial Engineering and Engineering Management
  (IEEM 2017)}.\hskip 1em plus 0.5em minus 0.4em\relax IEEE, 2017, pp.
  1357--1361.

\bibitem{lu2017adaptable}
Q.~Lu and X.~Xu, ``Adaptable blockchain-based systems: a case study for product
  traceability,'' \emph{IEEE Software}, vol.~34, no.~6, pp. 21--27, 2017.

\bibitem{toyoda2017novel}
K.~Toyoda, P.~T. Mathiopoulos, I.~Sasase, and T.~Ohtsuki, ``A novel
  blockchain-based product ownership management system (poms) for
  anti-counterfeits in the post supply chain,'' \emph{IEEE Access}, vol.~5, pp.
  17\,465--17\,477, 2017.

\bibitem{christidis2016blockchains}
K.~Christidis and M.~Devetsikiotis, ``Blockchains and smart contracts for the
  internet of things,'' \emph{{IEEE Access}}, vol.~4, pp. 2292--2303, 2016.

\bibitem{scekic2018blockchain}
O.~Scekic, S.~Nastic, and S.~Dustdar, ``Blockchain-supported smart city
  platform for social value co-creation and exchange,'' \emph{IEEE Internet
  Computing}, vol.~23, no.~1, pp. 19--28, 2019.

\bibitem{seebacher2017blockchain}
S.~Seebacher and R.~Sch{\"u}ritz, ``Blockchain technology as an enabler of
  service systems: A structured literature review,'' in \emph{International
  Conference on Exploring Services Science}, 2017, pp. 12--23.

\bibitem{kondrateva2021potential}
G.~Kondrateva, E.~de~Boissieu, C.~Ammi, and E.~Seulliet, ``The potential use of
  blockchain technology in co-creation ecosystems,'' \emph{Journal of
  Innovation Economics Management}, pp. I104--27, 2021.

\bibitem{nakamoto2008bitcoin}
S.~Nakamoto, ``Bitcoin: A peer-to-peer electronic cash system,''
  \emph{Preprint. \url{https://bitcoin.org/bitcoin.pdf}}, 2008.

\bibitem{ide2019efficient}
T.~Id{\'e}, R.~Raymond, and D.~T. Phan, ``Efficient protocol for collaborative
  dictionary learning in decentralized networks.'' in \emph{Proceeding of the
  28th International Joint Conference on Artificial Intelligence (IJCAI-19)},
  2019, pp. 2585--2591.

\bibitem{wu2005agreement}
C.~W. Wu, ``Agreement and consensus problems in groups of autonomous agents
  with linear dynamics,'' in \emph{Proceedings of the 2005 IEEE International
  Symposium on Circuits and Systems (ISCAS)}.\hskip 1em plus 0.5em minus
  0.4em\relax IEEE, 2005, pp. 292--295.

\bibitem{ren2005survey}
W.~Ren, R.~W. Beard, and E.~M. Atkins, ``A survey of consensus problems in
  multi-agent coordination,'' in \emph{Proceedings of the 2005 American Control
  Conference}.\hskip 1em plus 0.5em minus 0.4em\relax IEEE, 2005, pp.
  1859--1864.

\bibitem{olfati2007consensus}
R.~Olfati-Saber, J.~A. Fax, and R.~M. Murray, ``Consensus and cooperation in
  networked multi-agent systems,'' \emph{Proceedings of the IEEE}, vol.~95,
  no.~1, pp. 215--233, 2007.

\bibitem{vadhan2012pseudorandomness}
S.~P. Vadhan \emph{et~al.}, ``Pseudorandomness,'' \emph{Foundations and Trends
  in Theoretical Computer Science}, vol.~7, no. 1--3, pp. 1--336, 2012.

\bibitem{lubotzky2010discrete}
A.~Lubotzky, \emph{Discrete groups, expanding graphs and invariant
  measures}.\hskip 1em plus 0.5em minus 0.4em\relax Springer Science \&
  Business Media, 2010.

\bibitem{mohassel2017secureml}
P.~Mohassel and Y.~Zhang, ``{SecureML}: A system for scalable
  privacy-preserving machine learning,'' in \emph{Proceedings of the 2017 IEEE
  Symposium on Security and Privacy (SP)}.\hskip 1em plus 0.5em minus
  0.4em\relax IEEE, 2017, pp. 19--38.

\bibitem{mcmahan2017communication}
B.~McMahan, E.~Moore, D.~Ramage, S.~Hampson, and B.~A.~y. Arcas,
  ``Communication-efficient learning of deep networks from decentralized
  data,'' in \emph{Proceedings of the 20th International Conference on
  Artificial Intelligence and Statistics (AISTATS)}, 2017, pp. 1273--1282.

\bibitem{Konecny16NIPSworkshop}
J.~Kone\v{c}n\'{y}, H.~B. McMahan, F.~X. Yu, P.~Richt\'{a}rik, A.~T. Suresh,
  and D.~Bacon, ``Federated learning: Strategies for improving communication
  efficiency,'' in \emph{NIPS Workshop on Private Multi-Party Machine
  Learning}, 2016.

\bibitem{agarwal2018cpsgd}
N.~Agarwal, A.~T. Suresh, F.~Yu, S.~Kumar, and H.~B. McMahan, ``{cpSGD}:
  Communication-efficient and differentially-private distributed {SGD},'' in
  \emph{Advances in Neural Information Processing Systems}, 2018, pp.
  7575--7586.

\bibitem{yang2019federated}
Q.~Yang, Y.~Liu, T.~Chen, and Y.~Tong, ``Federated machine learning: Concept
  and applications,'' \emph{ACM Transactions on Intelligent Systems and
  Technology (TIST)}, vol.~10, no.~2, p.~12, 2019.

\bibitem{xu2014distributed}
M.~Xu, B.~Lakshminarayanan, Y.~W. Teh, J.~Zhu, and B.~Zhang, ``Distributed
  {Bayesian} posterior sampling via moment sharing,'' in \emph{Advances in
  Neural Information Processing Systems}, 2014, pp. 3356--3364.

\bibitem{Mohri19ICML}
M.~Mohri, G.~Sivek, and A.~T. Suresh, ``Agnostic federated learning,'' in
  \emph{Proceedings of the 36th International Conference on Machine Learning
  (ICML 2019)}, 2019, pp. 4615--4625.

\bibitem{smith2017federated}
V.~Smith, C.-K. Chiang, M.~Sanjabi, and A.~S. Talwalkar, ``Federated multi-task
  learning,'' in \emph{Advances in Neural Information Processing Systems},
  2017, pp. 4424--4434.

\bibitem{bernstein2018differentially}
G.~Bernstein and D.~R. Sheldon, ``Differentially private {Bayesian} inference
  for exponential families,'' in \emph{Advances in Neural Information
  Processing Systems}, 2018, pp. 2924--2934.

\bibitem{xie2017privacy}
L.~Xie, I.~M. Baytas, K.~Lin, and J.~Zhou, ``Privacy-preserving distributed
  multi-task learning with asynchronous updates,'' in \emph{Proceedings of the
  23rd ACM SIGKDD International Conference on Knowledge Discovery and Data
  Mining}.\hskip 1em plus 0.5em minus 0.4em\relax ACM, 2017, pp. 1195--1204.

\bibitem{heikkila2017differentially}
M.~Heikkil{\"a}, E.~Lagerspetz, S.~Kaski, K.~Shimizu, S.~Tarkoma, and
  A.~Honkela, ``Differentially private {Bayesian} learning on distributed
  data,'' in \emph{Advances in neural information processing systems}, 2017,
  pp. 3226--3235.

\bibitem{ding2018comparing}
B.~Ding, H.~Nori, P.~Li, and J.~Allen, ``Comparing population means under local
  differential privacy: with significance and power,'' in \emph{Proceedings of
  the Thirty-Second AAAI Conference on Artificial Intelligence (AAAI-18)},
  2018.

\bibitem{liu2020secure}
Y.~Liu, Y.~Kang, C.~Xing, T.~Chen, and Q.~Yang, ``A secure federated transfer
  learning framework,'' \emph{IEEE Intelligent Systems}, vol.~35, no.~4, pp.
  70--82, 2020.

\bibitem{ruan2017secure}
M.~Ruan, M.~Ahmad, and Y.~Wang, ``Secure and privacy-preserving average
  consensus,'' in \emph{Proceedings of the 2017 Workshop on Cyber-Physical
  Systems Security and Privacy}.\hskip 1em plus 0.5em minus 0.4em\relax ACM,
  2017, pp. 123--129.

\bibitem{ide2018collaborative}
T.~Id{\'e}, ``Collaborative anomaly detection on blockchain from noisy sensor
  data,'' in \emph{Proceeding of the 2018 IEEE International Conference on Data
  Mining Workshops (ICDMW)}.\hskip 1em plus 0.5em minus 0.4em\relax IEEE, 2018,
  pp. 120--127.

\bibitem{betulehmann2006theory}
E.~L. Lehmann and G.~Casella, \emph{Theory of point estimation}.\hskip 1em plus
  0.5em minus 0.4em\relax Springer Science \& Business Media, 2006.

\bibitem{Friedman08glasso}
J.~Friedman, T.~Hastie, and R.~Tibshirani, ``Sparse inverse covariance
  estimation with the graphical lasso,'' \emph{Biostatistics}, vol.~9, no.~3,
  pp. 432--441, 2008.

\bibitem{hsieh2014quic}
C.-J. Hsieh, M.~A. Sustik, I.~S. Dhillon, and P.~Ravikumar, ``{QUIC}: quadratic
  approximation for sparse inverse covariance estimation,'' \emph{Journal of
  Machine Learning Research}, vol.~15, no.~1, pp. 2911--2947, 2014.

\bibitem{Strang1976}
G.~Strang, \emph{Linear Algebra and its Applications}.\hskip 1em plus 0.5em
  minus 0.4em\relax Academic Press, 1976.

\bibitem{shamir1979share}
A.~Shamir, ``How to share a secret,'' \emph{Communications of the ACM},
  vol.~22, no.~11, pp. 612--613, 1979.

\bibitem{boneh2016graduate}
D.~Boneh and V.~Shoup, ``A graduate course in applied cryptography,''
  \emph{Draft of a book, version 0.3, December}, 2016.

\bibitem{xiao2020survey}
Y.~Xiao, N.~Zhang, W.~Lou, and Y.~T. Hou, ``A survey of distributed consensus
  protocols for blockchain networks,'' \emph{IEEE Communications Surveys \&
  Tutorials}, vol.~22, no.~2, pp. 1432--1465, 2020.

\bibitem{karlof2003secure}
C.~Karlof and D.~Wagner, ``Secure routing in wireless sensor networks: Attacks
  and countermeasures,'' in \emph{Proceedings of the First IEEE International
  Workshop on Sensor Network Protocols and Applications, 2003.}\hskip 1em plus
  0.5em minus 0.4em\relax IEEE, 2003, pp. 113--127.

\bibitem{gupta2017privacy}
N.~Gupta, J.~Katz, and N.~Chopra, ``Privacy in distributed average consensus,''
  \emph{IFAC-PapersOnLine}, vol.~50, no.~1, pp. 9515--9520, 2017.

\bibitem{davidoff2003elementary}
G.~Davidoff, P.~Sarnak, and A.~Valette, \emph{Elementary number theory, group
  theory and Ramanujan graphs}.\hskip 1em plus 0.5em minus 0.4em\relax
  Cambridge University Press, 2003.

\bibitem{brouwer2011spectra}
A.~E. Brouwer and W.~H. Haemers, \emph{Spectra of graphs}.\hskip 1em plus 0.5em
  minus 0.4em\relax Springer Science \& Business Media, 2011.

\bibitem{Friedman2004A}
J.~Friedman, ``A proof of {Alon}'s second eigenvalue conjecture,'' in
  \emph{Proceedings of the Thirty-Fifth Annual ACM Symposium on Theory of
  Computing (STOC '03)}, 2003, p. 720–724.

\bibitem{kim2006generating}
J.~Kim and V.~Vu, ``Generating random regular graphs,'' \emph{Combinatorica},
  vol.~26, no.~6, pp. 683--708, 2006.

\bibitem{homomorpheR}
B.~Narasimhan, ``{homomorpheR},'' in \emph{CRAN (The Comprehensive \texttt{R}
  Archive Network)}, 2019.

\bibitem{RuanGaoWang2019}
M.~Ruan, H.~Gao, and Y.~Wang, ``Secure and privacy-preserving consensus,''
  \emph{IEEE Transactions on Automatic Control}, 2019.

\bibitem{cheon2017homomorphic}
J.~H. Cheon, A.~Kim, M.~Kim, and Y.~Song, ``Homomorphic encryption for
  arithmetic of approximate numbers,'' in \emph{Proc. the 2017 Intl. Conf. the
  Theory and Application of Cryptology and Information Security (ASIACRYPT
  2017)}.\hskip 1em plus 0.5em minus 0.4em\relax Springer, 2017, pp. 409--437.

\bibitem{minami2016differential}
K.~Minami, H.~Arai, I.~Sato, and H.~Nakagawa, ``Differential privacy without
  sensitivity,'' in \emph{Advances in Neural Information Processing Systems},
  2016, pp. 956--964.

\end{thebibliography}
